\documentclass[10pt,twocolumn,letterpaper]{article}
\usepackage[accsupp]{axessibility}
\usepackage{cvpr}              

\usepackage{booktabs}

\usepackage{amsmath,amsfonts,bm}

\def\1{\bm{1}}

\DeclareMathAlphabet{\mathsfit}{\encodingdefault}{\sfdefault}{m}{sl}
\SetMathAlphabet{\mathsfit}{bold}{\encodingdefault}{\sfdefault}{bx}{n}

\usepackage{url} 

\usepackage[utf8]{inputenc} 
\usepackage{cite}
\usepackage[T1]{fontenc}    
\usepackage{nicefrac}       
\usepackage{microtype}      
\usepackage{xcolor} 
\usepackage{graphicx}
\usepackage{wrapfig}
\usepackage{amsmath,amssymb,amsfonts}
\usepackage{threeparttable}

\usepackage[pagebackref,breaklinks,colorlinks]{hyperref}

\usepackage[capitalize]{cleveref}
\crefname{section}{Sec.}{Secs.}
\Crefname{section}{Section}{Sections}
\Crefname{table}{Table}{Tables}
\crefname{table}{Tab.}{Tabs.}

\graphicspath{{ ./figures/}}
\usepackage{adjustbox}

\usepackage{subcaption}

\def\cvprPaperID{11327} 
\def\confName{CVPR}
\def\confYear{2022}

\def\A{{\bf A}}

\def\D{{\bf D}}

\def\X{{\bf X}}

\def\x{{\bf x}}

\def\z{{\bf z}}
\def\Z{{\bf Z}}

\def\0{{\bf 0}}
\def\1{{\bf 1}}

\newtheorem{theorem}{Theorem}

\numberwithin{theorem}{section}
\numberwithin{lemma}{section}
\numberwithin{remark}{section}
\numberwithin{cor}{section}
\numberwithin{proposition}{section}

\newcommand{\secref}[1]{Section~\ref{#1}}
\newcommand{\figref}[1]{Figure~\ref{#1}}

\newcommand{\eqnref}[1]{Eqn.~\ref{#1}}

\begin{document}
\title{{\em OrphicX}: A Causality-Inspired Latent Variable Model for Interpreting Graph Neural Networks}

\author{
Wanyu Lin$^1$
\and
Hao Lan$^2$
\and Hao Wang$^3$
\and Baochun Li$^2$
\\
$^1$The Hong Kong Polytechnic University, $^2$University of Toronto, $^3$Rutgers University\\
{\tt\small wan-yu.lin@polyu.edu.hk, hao.lan@mail.utoronto.ca, bli@ece.toronto.edu, hoguewang@gmail.com}}

\maketitle

\begin{abstract}
   This paper proposes a new e{\em X}planation framework, called {\em OrphicX}, for generating causal explanations for any graph neural networks (GNNs) based on learned {\em latent} causal factors. Specifically, we construct a distinct generative model and design an objective function that encourages the generative model to produce causal, compact, and faithful explanations. This is achieved by isolating the causal factors in the latent space of graphs by maximizing the information flow measurements. We theoretically analyze the cause-effect relationships in the proposed causal graph, identify node attributes as confounders between graphs and GNN predictions, and circumvent such confounder effect by leveraging the backdoor adjustment formula. Our framework is compatible with any GNNs, and it does not require access to the process by which the target GNN produces its predictions. In addition, it does not rely on the linear-independence assumption of the explained features, nor require prior knowledge on the graph learning tasks. We show a proof-of-concept of {\em OrphicX} on canonical classification problems on graph data. In particular, we analyze the explanatory subgraphs obtained from explanations for molecular graphs (i.e., Mutag) and quantitatively evaluate the explanation performance with frequently occurring subgraph patterns. Empirically, we show that {\em OrphicX} can effectively identify the causal semantics for generating causal explanations, significantly outperforming its alternatives. 
\end{abstract}


\section{Introduction}
\label{sec:intro}


Graph neural networks (GNNs) have found various applications in many scientific domains, including iamge classification~\cite{lin2020shoestring}, 3D-shape analysis~\cite{monti2017geometric}, video analysis~\cite{YuanyuanICCV2017}, speech recognition~\cite{DGP}, 
and social information systems~\cite{wanyu-infocom21,wanyu-infocom20}. The decisions of powerful GNNs for graph-structural data are difficult to interpret. In this paper, we focus on providing post-hoc explanations for any GNN by parameterizing the process of generating explanations. Specifically, given a pre-trained GNN of interest, an explanation model, or called {\em explainer}, is trained for generating compact subgraphs, leading to the model outcomes. However, learning the explanation process can be difficult as no ground-truth explanations exist. If an explanation highlights subjectively irrelevant subgraph patterns of the input instance, this may correctly reflect the target GNN’s unexpected way of processing the data, or the explanation may be inaccurate.

Recently, a few recent works have been proposed to explain GNNs via learning the explanation process. XGNN~\cite{yuan2020xgnn} was proposed to investigate the graph patterns that lead to a specific class by learning a policy network. PGExplainer~\cite{luo2020parameterized} was proposed to learn a mask predictor to obtain the edge masks for providing explanations. However, XGNN fails to explain individual instances and therefore lacks local fidelity~\cite{ribeiro2016why}, while PGExplainer heavily relies on the learned embeddings of the target model, and has the restrictive assumption of having domain knowledge over the learning tasks (e.g., the explicit subgraph patterns are provided). The closest to ours is Gem~\cite{wanyuicml21}, wherein an explainer is learned based on the concept of Granger causality. The distillation process of ground-truth explanation naturally implies the independent assumptions of the explained features\footnote{We are aware of the drawbacks of reusing the term ``feature.'' Specifically, nodes and edges are the explained features in an explanatory subgraph.}, which might be problematic as the graph-structured data is inherently interdependent~\cite{GRDA}. 

In this work, we define a distinct generative model as an explainer that can provide interpretable explanations for any GNNs through the lens of causality, in particular from the notion of the structural causal model (SCM)~\cite{pearl2009causality,ICL}. In principle, generating causal explanations require reasoning about how changing different concepts of the input instance --- which can be thought of as enforcing perturbations or interventions on the input --- affects the decisions over the target model (or the response of the system)~\cite{GenInt}. Different from prior works quantifying the causal influence from the data space (e.g., Gem~\cite{wanyuicml21}), we propose to identify the underlying causal factors from latent space. By doing so, we can avoid working with input spaces with complex interdependency. The intuition is that if the latent features\footnote{Features and factors are used interchangeably, e.g., causal features are equivalent to causal factors.} can untwist the causal factors and the spurious factors between the input instance and the corresponding output of the target GNN, generating causal explanations is possible.

For this purpose, we first present a causal graph that models both causal features and spurious features to the GNN's prediction. The causal features causing the prediction might be informative to generate a graph-structural mask for the explanation. Our causal analysis shows that there exists a confounder from the data space while considering the cause-effect relationships between the latent features and the GNN outcome~\cite{ExposureBias,RDL,OEM}. Specifically, when interpreting graph-structural data, node features/attributes can be a confounder that affects both the generated graph structures and corresponding model outcomes. The existence of the confounder represents a barrier to causal quantification~\cite{pearl2009causality}. To this end, we adopt the concept of information flow~\cite{informationflow2008}, along with the backdoor adjustment formula~\cite{CausalPrimer}, to bypass the confounder effect and measure the causal information transmission from the latent features to the predictions.

\begin{figure*}[t]
\vskip -0.4cm
  \centering
  \includegraphics[scale = 0.145]{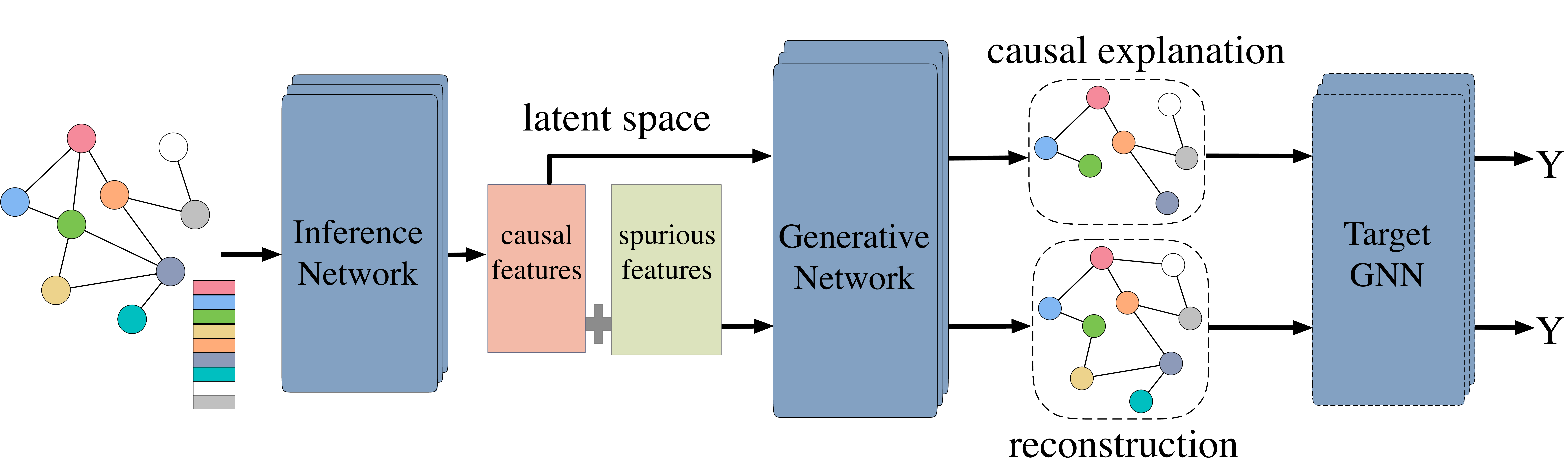}
  \vskip -0.3cm
  \caption{Illustration of {\em OrphicX}. We instantiate our explainer with a variational graph auto-encoder (VGAE), which consists of an inference network and a generative network. The causal features along with the spurious features can be used to reconstruct the graph structure within the data distribution, while the causal features are mapped to a graph-structured mask for the causal explanation. The target GNN is pre-trained, and the parameters would not be changed during the training of {\em OrphicX}.}
  \vspace{-10pt}
  \label{fig:arc}
\end{figure*}

Then we instantiate our explainer with a variational graph auto-encoder (VGAE)~\cite{kipf2016variational}, which consists of an inference network and a generative network (shown in Figure~\ref{fig:arc}). The inference network seeks a representation of the input, in which the representation is learned in such a way that a subset of the factors with large causal influence, i.e. the causal features, can be identified. The generative network is to map the causal features into an adjacency mask for the explanation. Importantly, the generative network ensures that the learned latent representations (the causal features and the spurious features together) are within the data distribution.

In a nutshell, our main contributions are highlighted as follows. We propose a new explanation technique, called {\em OrphicX}, that eXplains the predictions of any GNN by identifying the causal factors in the {\em latent} space. We utilize the notion of information flow measurements to quantify the causal information flowing from the latent features to the model predictions. We theoretically analyze the causal-effect relationships in the proposed causal model, identify a confounder, and circumvent it by leveraging the backdoor adjustment formula. We empirically demonstrate that the learned features with causal semantics are indeed informative for generating interpretable and faithful explanations for any GNNs. Our work improves model interpretability and increases trust in GNN model explanation results.

\section{Method}
\label{sec:method}
\subsection{Notations and Problem Setting}
\label{sub:problem}

\textbf{Notations.} Given a pre-trained GNN (the target model to be explained), denoted as $f:\,\mathcal{G}\,\rightarrow\,\mathcal{Y}$, where $\mathcal{G}$ is the space of input graphs to the model and $\mathcal{Y}$ is the label space. Specifically, the input graph $G=(V, E)$ of the GNN includes the corresponding adjacency matrix ($\A\in\mathbb{R}^{ |V|\times|V|}$) and a node attribute matrix ($\X\in \mathbb{R}^{|V|\times D}$). We use $\Z = [\Z_c, \Z_s] \in \mathbb{R}^{|V|\times (D_c+D_s)}$ to denote the latent feature matrix, where $\Z_c$ is the causal feature sub-matrix and $\Z_s$ is the spurious feature sub-matrix. Correspondingly for each node, we denote its node attribute vector by $\x$ (one row of $\X$), its causal latent features by $\z_c$, and its spurious latent features by $\z_s$. 

\textbf{The desiderata for GNN explanation methods.} An essential criterion for explanations is {\em fidelity}~\cite{ribeiro2016why}. A faithful explanation/subgraph should correspond to how the target GNN behaves in the vicinity of the given graph of interest. Stated differently, the outcome of feeding to the explanatory subgraph to the target GNN should be similar to that of the graph to be explained. Another essential criterion for explanations is human interpretability, which implies that the generated explanations should be {\em sparse/compact} in the context of graph-structured data~\cite{pope2019explainability}. In other words, a human-understandable explanation should highlight the most important part of the input while discarding the irrelevant part. In addition, an explainer should be able to explain any GNN model, commonly known as  {\em ``model-agnostic''} (i.e., treat the target GNN as a black box). 

\begin{figure}[t]
  \centering
  \includegraphics[scale = 0.15]{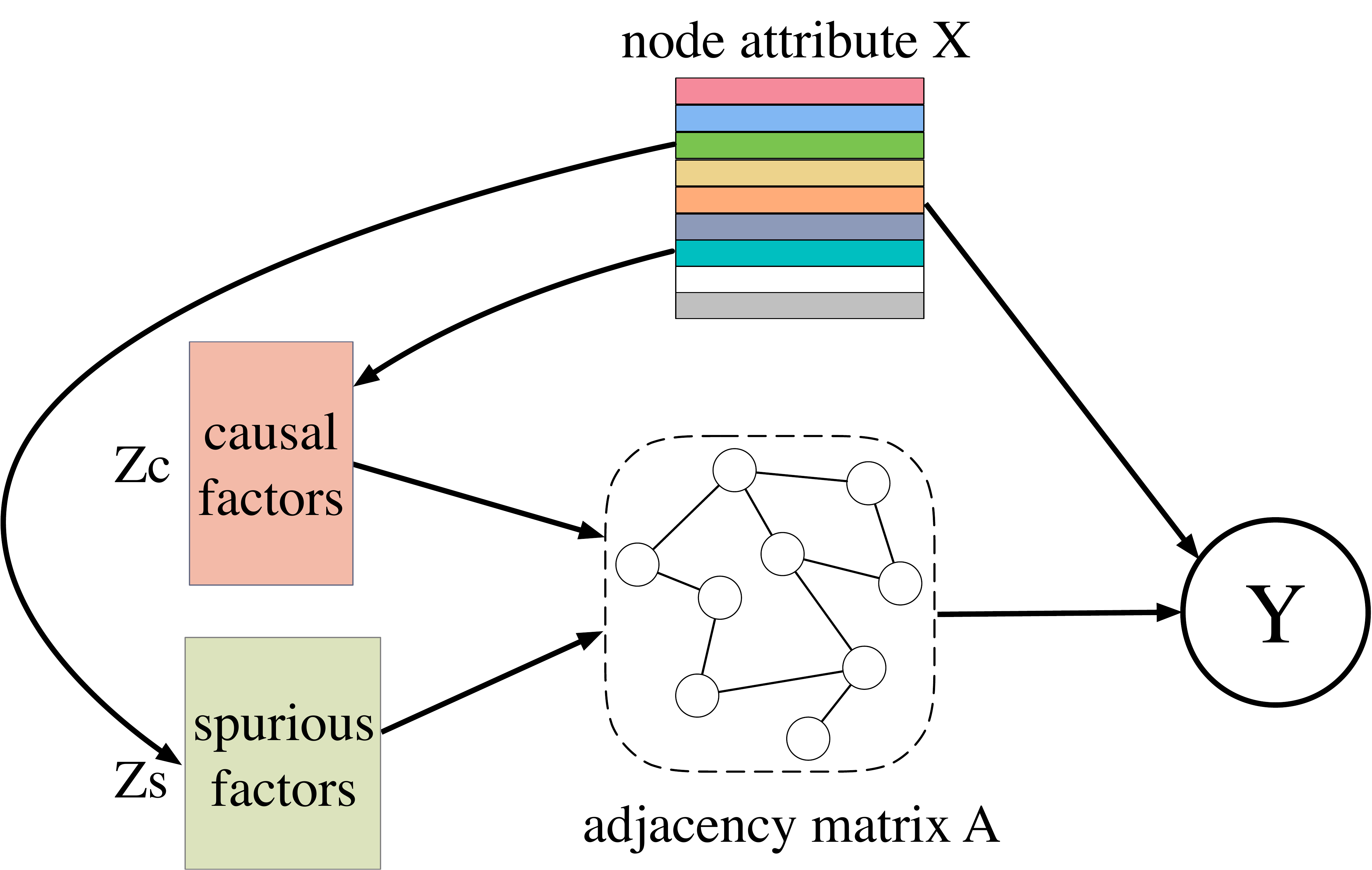}
  \caption{Illustration of the causal graph. The causal features are a set of factors in the latent space. The causal features and the spurious features together form the representation of the input graph. The graph structure is reconstructed based on the latent representation; it forms the input of the target GNN, along with the feature matrix. $y$ denotes the predicted label of the GNN target.}
  \vspace{-10pt}
  \label{fig:scm}
\end{figure}

\textbf{Problem setting.} Therefore, our ultimate goal is to obtain a generative model as an explainer, denoted as $\mathcal{F}$, that can identify which part of the input causes the GNN prediction, while achieving the best possible performance under the above criteria. Consistent with prior works~\cite{yuan2020xgnn,wanyuicml21,luo2020parameterized}, we focus on explanations on graph structures. We consider the black-box setting where we do not have any information about the ground-truth labels of the input graphs and we specifically do not require access to, nor knowledge of, the process by which the target GNN produces its output. Nevertheless, we are allowed to retrieve different predictions by performing multiple queries, and we assume that the gradients of the target GNN are provided. 

\subsection{{\em OrphicX}}

\textbf{Overview.}
In this paper, we propose a generative model as an explainer, called {\em OrphicX}, that can generate causal explanations by identifying the causal features leading to the GNN outcome. In particular, we propose to isolate the causal features and the spurious features from the latent space. For this purpose, we first propose a causal graph to model the relationships among the causal features, the spurious features, the input graph, and the prediction of the target model. Then we show how to train {\em OrphicX} with a faithful causal-quantification mechanism based on the notion of information flow along with the backdoor adjustment formula. With the identified causal features, we are able to generate a graph-structured mask for the explanation.

\textbf{Information flow for causal measurements.} 
Recall that, our objective is to generate compact subgraphs as the explanations for the pre-trained GNN. The explanatory subgraph is causal in the sense that it tends to be independent of the spurious aspects of the input graph while holding the causal portions contributing to the prediction of the target GNN. One challenge, therefore, is how to quantify the causal influence of different data aspects in the latent space, so as to identify the portion with large causal influence, denoted by $\mathbf{Z}_c$. To address this issue, we leverage recent work on information-theoretic measures of causal influence~\cite{informationflow2008}. Specifically, we measure the causal influence of $\Z_c$ on the model prediction $y$ using the information flow, denoted as $\mathbf{I}\left(\mathbf{Z}_c\rightarrow y\right)$, between them. Here information flow can be seen as the causal counterpart of mutual information $\mathbf{I}\left(\mathbf{Z}_c; y\right)$. 

Succinctly, our framework attempts to isolate a subset of the representation from the hidden space, denoted as $\mathbf{Z}_c$, such that the information flow from $\mathbf{Z}_c$ to $y$ is maximized. In what follows, we will show how to quantify this term corresponding to our causal model.

\textbf{Causal analysis.} Throughout this paper, we assume the causal model in Figure~\ref{fig:scm}. Specifically, the causal features and the spurious features together form the representation of the input graph, which can be used to reconstruct the graph structure, denoted as $\A$. This ensures that the learned latent features still reflect the same data distribution as the one captured by the target GNN. The graph structure $\A$, along with the node attribute $\X$, contributes to the model prediction $y$. Stated differently, $\X$ is a confounder when we consider the cause-effect relationships between the latent features (i.e. causal features and spurious features) and the model prediction. Consequently, directly ignoring $\X$ can lead to inaccurate estimates of the causal features. To address this issue, we leverage the classic backdoor adjustment formula~\cite{CausalPrimer} and have:
\begin{align}
    P(y | do(\Z_c)) = \sum_{\X} P(y | \Z_c, \X) P(\X).\label{eq:backdoor}
\end{align}
\eqnref{eq:backdoor} is crucial to circumvent the confounder effect introduced by node attributes and compute the information flow $I(\Z_c\rightarrow y)$, which is the causal counterpart of mutual information~\cite{informationflow2008}. Intuitively, \eqnref{eq:backdoor} goes through different versions of $\X$ while keeping $\Z_c$ fixed to estimate the causal effect $\Z_c$ has on $y$. Note that $P(y | do(\Z_c))=\sum_{\X}P(y | \Z_c, \X) P(\X)$ is different from $P(y | \Z_c) =  \sum_{\X}P(y | \Z_c, \X) P(\X|\Z_c)$; the former samples from the marginal distribution $P(\X)$, while the latter samples $\X$ from the conditional distribution $P(\X|\Z_c)$. In causal theory, $P(y | do(\Z_c))=\sum_{\X}P(y | \Z_c, \X) P(\X)$ is referred to as the backdoor adjustment formula~\cite{CausalPrimer}. Our Theorem~\ref{thm:info_flow} below provides a way of computing the information flow $I(\Z_c\rightarrow y)$.

\begin{theorem}[\textbf{Information flow between $\Z_c$ and $y$}]\label{thm:info_flow}
The information flow between the causal factors $\Z_c$ and the prediction $y$ can be computed as
\begingroup\makeatletter\def\f@size{8}\check@mathfonts
\begin{align*}
&I(\Z_c\rightarrow y) \\
= &\int_{\Z_c} P(\Z_c) \sum\limits_y P(y | do(\Z_c)) 
\log \frac{P(y | do(\Z_c))}{\int_{\Z_c} P(y | do(\Z_c)) d\Z_c} d\Z_c\\
= &\int_{\Z_c} P(\Z_c) \sum\limits_y \sum_{\X} P(y | \Z_c, \X) P(\X) \cdot\\
&\quad\quad\log \frac{\sum_{\X} P(y | \Z_c, \X) P(\X)}{\int_{\Z_c} \sum_{\X} P(y | \Z_c, \X) P(\X) d\Z_c} d\Z_c
\end{align*}
\endgroup
\end{theorem}

Note that due to the confounder $\X$, $I(\Z_c\rightarrow y)$ is \emph{not} equal to the mutual information $I(\Z_c; y)$. The term $\sum_{\X} P(y | \Z_c, \X)$ comes from \eqnref{eq:backdoor} and can be estimated efficiently. Specifically, we have
\begingroup\makeatletter\def\f@size{8}\check@mathfonts
\begin{align}
&P(y|do(\Z_c)) 
=\sum_{\X} P(y | \Z_c, \X) P(\X)\\
&=\sum_{\X} \sum_{\A} \int_{\Z_s} P(y|\A,\X) P(\A|\Z_s,\Z_c) P(\Z_s | \Z_c, \X) P(\X)d\Z_s\nonumber\\
&\approx \frac{1}{N_x N_s N_z} \sum\limits_{k=1}^{N_x}\sum\limits_{j=1}^{N_s}\sum\limits_{n=1}^{N_z}P(y|\A^{(kjn)}, \X^{(k)}).\label{eq:yzc}
\end{align}
\endgroup
Here $k$ indexes the $N_x$ sampled node attribute matrices $\X^{(k)}$ from the dataset; 
$j$ indexes the $N_s$ samples for each $\X^{(k)}$, i.e., $\Z_s^{(kj)}\sim P(\Z_s | \Z_c, \X^{(k)})$; 
$n$ indexes the $N_z$ sampled graphs for each $\Z_s^{(kj)}$, i.e., $\A^{(kjn)}\sim p(\A|\Z_c, \Z_s^{(kj)})$. 
Note that in practice we use the variational distribution $q(\Z_s|\A, \X^{(k)})$ to approximate the true posterior distribution $P(\Z_s | \Z_c, \X^{(k)})$, and that in~\eqnref{eq:yzc}, $\X$, $\Z_c$, and $\Z_s$ do not necessarily belong to the same graph in the original dataset. Intuitively this is to remove the confounding effect of $\X$ on $\Z_c$ and $\Z_s$. Consequently we have
\vspace{-10pt}

\begingroup\makeatletter\def\f@size{10}\check@mathfonts
\begin{align}
&\int_{\Z_c} P(\Z_c) P(y|do(\Z_c)) d\Z_c \\
&=\int_{\Z_c}\sum_{\X} \sum_{\A} \int_{\Z_s} P(y|\A,\X) P(\A|\Z_s,\Z_c) \\
&\quad\quad\quad\quad\quad\quad\quad\quad P(\Z_s | \Z_c, \X) P(\X) P(\Z_c)d\Z_s d\Z_c\nonumber\\
&\approx \frac{1}{N_c N_x N_s N_z} \sum\limits_{i=1}^{N_c} \sum\limits_{k=1}^{N_x}\sum\limits_{j=1}^{N_s}\sum\limits_{n=1}^{N_z}P(y|\A^{(ikjn)}, \X^{(k)}),\label{eq:zcyzc}
\end{align}
\endgroup
Similarly, 
$i$ indexes the $N_c$ samples from $\Z_c$'s marginal distribution, i.e., $\Z_c^{(i)}\sim P(\Z_c)$; 
$k$ indexes the $N_x$ sampled node attribute matrices from $\X$'s marginal distribution $\X^{(k)}\sim P(\X)$; 
$j$ indexes the $N_s$ samples of $\Z_s$ for each pair $(\Z_c^{(i)},\X^{(k)})$, i.e., $\Z_s^{(ikj)}\sim P(\Z_s|\Z_c^{(i)}, \X^{(k)})$;
$n$ indexes the $N_z$ sampled graphs for each pair $(\Z_c^{(i)}, \Z_s^{(kj)})$, i.e., $\A^{(ikjn)}\sim p(\A|\Z_c^{(i)}, \Z_s^{(kj)})$. 
Note that in practice we use the variational distribution $q(\Z_s|\A, \Z_c^{(i)}, \X^{(k)})$ to approximate the true posterior distribution $P(\Z_s|\Z_c^{(i)}, \X^{(k)})$.

Put together, we have
\vspace{-7pt}

\begingroup\makeatletter\def\f@size{7.3}\check@mathfonts
\begin{align*}
I(\Z_c\rightarrow y) = & \frac{1}{N_c N_x N_s N_z} \Big[ \sum_{i=1}^{N_c} \sum_y \Big( \sum\limits_{k=1}^{N_x}\sum\limits_{j=1}^{N_s}\sum\limits_{n=1}^{N_z}P(y|\A^{(ikjn)}, \X^{(k)})\Big)\cdot\\
&\log \Big(\frac{1}{N_x N_s N_z} \sum\limits_{k=1}^{N_x}\sum\limits_{j=1}^{N_s}\sum\limits_{n=1}^{N_z}P(y|\A^{(ikjn)}, \X^{(k)})\Big) \\
&- \sum_y \Big( \sum\limits_{i=1}^{N_c} \sum\limits_{k=1}^{N_x}\sum\limits_{j=1}^{N_s}\sum\limits_{n=1}^{N_z}P(y|\A^{(ikjn)}, \X^{(k)})\Big)\cdot\\
& \log \Big(\frac{1}{N_c N_x N_s N_z} \sum\limits_{i=1}^{N_c} \sum\limits_{k=1}^{N_x}\sum\limits_{j=1}^{N_s}\sum\limits_{n=1}^{N_z}P(y|\A^{(ikjn)}, \X^{(k)})\Big) \Big].
\end{align*}
\endgroup

{\bf Graph generative model as an explainer.}
Our framework, {\em OrphicX}, leverages the latent space of a variational graph auto-encoder (VGAE) to avoid working with input spaces with complex interdependency. Specifically, our VGAE-based framework (shown in Figure~\ref{fig:arc}) consists of an inference network and a generative network. The former is instantiated with a graph convolutional encoder and the latter is a multi-layer perceptron equipped with an inner product decoder. More concretely, the inference network seeks a representation --- a latent feature matrix $\Z$ of the input graph, of which the causal features $\Z_c$, a sub-matrix with large causal influence, can be isolated. The generative network serves two purposes: (1) it maps the causal sub-matrix into an adjacency mask, which is used as the causal explanation, and (2) it ensures that the causal features, merged with the spurious features, can reconstruct the graphs within the data distribution characterized by the target GNN. 

\textbf{Learning {\em OrphicX}.} Learning of {\em OrphicX} can be cast as the following optimization problem: 
\vspace{-5pt}
\begin{equation}\label{eq:OrphicX_causal}
\min\,\,-{I}\left(\mathbf{Z}_c\rightarrow y\right) + \lambda\mathcal{L}_{\mathbf{VGAE}},
\end{equation}

where $\mathcal{L}_{\mathbf{VGAE}}$ is the negative evidence lower bound (ELBO) loss term that encourages the latent features $\Z$ to stay in the data manifold~\cite{kipf2016variational}, and $\mathbf{Z}_c$ is the causal sub-matrix of $\Z$. A detailed description of the ELBO term of the VGAE is provided in Appendix. Our empirical results suggest that the ELBO term helps learn a sub-matrix that embeds more relevant information leading to the GNN prediction. 

Recall that, our objective is to generate explanations that can provide insights into how the target GNN truly computes its predictions.  An ideal explainer should fulfill the three desiderata presented in~\secref{sub:problem}: high fidelity (faithful), high sparsity (compact), and model agnostic. Therefore, apart from the objective function~\eqnref{eq:OrphicX_causal}, we further enforce the fidelity and sparsity criteria through regularization specifically tailored to such explainers. Concretely, we denote the generated explanatory subgraph as $G_c$ and the corresponding adjacency matrix as $\A_c$. The sparsity criterion is measured by $\frac{||\A_c||_1}{||\A||_1}$, where $||\cdot||_1$ denotes the $l_1$ norm of the adjacency matrix. The fidelity criterion implies that the GNN outcome corresponding to the explanatory subgraph should be approximated to that of the target instance, i.e. $f(G_c)\approx f(G)$, where $f(\cdot)$ is the probability distribution over the classes --- the outcome of the target GNN. For this purpose, we introduce a Kullback–Leibler (KL) divergence term to measure how much the two outputs differ.

Therefore, the optimization problem can be reformulated as:
\vspace{-15pt}
\begin{align*}\label{eq:OrphicX}
\min\,-I\left(\mathbf{Z}_c\rightarrow y\right) + \lambda_1\mathcal{L}_{\mathbf{VGAE}} + \lambda_2 \frac{||\A_c||_1}{||\A||_1} \\
+ \lambda_3\mathbf{KL}\left(f(G_c),f(G)\right),
\end{align*}
where $\lambda_i$ ($i\in\{1,2,3\}$) controls the associated regularizer terms. To understand {\em OrphicX} comprehensively, a series of ablation studies for the loss function are performed. Note that, the parameters of the target GNN (shown in~\figref{fig:arc}) are pre-trained and would not be changed during the training of {\em OrphicX}. {\em OrphicX} only works with the model inputs and the outputs, rather than the internal structure of specific models. Therefore, our framework can be used to explain any GNN models as long as their gradients are admitted.

\section{Experiments}
\label{sec:experiments}
{\renewcommand{\arraystretch}{0.6}%

\subsection{Datasets and Settings}

{\bf Datasets.} We conducted experiments on benchmark datasets for interpreting GNNs: 1) For the node classification task, we evaluate different methods with synthetic datasets, including BA-shapes and Tree-cycles, where ground-truth explanations are available. We followed data processing in the literature~\cite{ying2019gnnexplainer}. 2) For the graph classification task, we use two datasets in bioinformatics, MUTAG~\cite{debnath1991structure} and NCI1~\cite{wale2008comparison}. Note that the model architectures for node classification~\cite{hamilton2017inductive} and graph classification~\cite{xu2018powerful} tasks are different (more details of the dataset descriptions and corresponding model architectures are provided in Appendix~\ref{appendix-implementation}).

{\bf Comparison methods.} We compare our approach against various powerful interpretability frameworks for GNNs. They are GNNExplainer~\cite{ying2019gnnexplainer}, PGExplainer~\cite{luo2020parameterized}, and Gem~\cite{wanyuicml21}\footnote{We use the source code released by the authors.}. Among others, PGExplainer and Gem explain the target GNN via learning an explainer. As for GNNExplainer, there is no training phase, as it is naturally designed for explaining a given instance at a time. Unless otherwise stated, we set all the hyperparameters of the baselines as reported in the corresponding papers. 

{\bf Hyperparameters in {\em OrphicX}.} For all datasets on different tasks, the explainers share the same model structure~\cite{kipf2016variational}. For the inference network, we applied a three-layer GCN with output dimensions $32$, $32$, and $16$. The generative model is equipped with a two-layer MLP and an inner product decoder. We trained the explainers using the Adam optimizer~\cite{kingma2014adam} with a learning rate of $0.003$ for $300$epochs. For all experiments, we set $N_x=5$, $N_z=2$, $N_c=25$, $N_s=100$, $D_c=3$, $\lambda_1=0.1$, $\lambda_2=0.1$, and $\lambda_3=0.2$. The results reported in the paper correspond to the best hyperparameter configurations. With this testing setup, our goal is to fairly compare best achievable explanation performance of the methods. Detailed implementations, including our hyperparameter search space are given in Appendix~\ref{appendix-implementation}.

{\bf Evaluation metrics.} We evaluate our approach with two criteria. 1) Faithfulness\footnote{In the context of model interpretability, ``faithfulness'' means high fidelity~\cite{lundberg2017unified}, which is different from the meaning used in causal discovery.}/fidelity: are the explanations indicative of ``true'' model behaviors? 2) Sparsity: are the explanations compact and understandable? Below, we address these criteria, proposing quantitative metrics for evaluating fidelity and sparsity and qualitative assessment via visualizing the explanations. 

To evaluate {\em fidelity}, we generate explanations for the test set\footnote{The detailed data splitting is provided in the Appendix.} according to {\em OrphicX}, Gem, PGExplainer, and GNNExplainer, respectively. We then evaluate the {\em explanation accuracy} of different methods by comparing the predicted labels of the explanatory subgraphs with the predicted labels of the input graphs using the pre-trained GNN~\cite{wanyuicml21}. An explanation is faithful only when the predicted label of the explanatory subgraph is the same as the corresponding input graph. To evaluate {\em sparsity}, we use different evaluation metrics. Specifically, in Mutag, the type and the size of explainable motifs are various. We measure the fraction of edges (i.e., edge ratio denoted as $R$) selected as ``important'' by different explanation methods for Mutag and NCI1. For the synthetic datasets, we use the number of edges (denoted as $K$), as did in prior works~\cite{wanyuicml21,ying2019gnnexplainer}. A smaller fraction of edges or a smaller number of edges selected implies a more compact subgraph or higher sparsity. 

To further check the {\em interpretability}, we use the visualized explanations to analyze the performance qualitatively. However, we do not know the ground-truth explanations for the real-world datasets. For Mutag\footnote{As we cannot obtain the ground-truth explanations for NCI1, we focus on the quantitative evaluation for this dataset.}, we ask an expert from Biochemical Engineering to label the explicit subgraph patterns as our explanation ground truth (i.e., carbon rings with chemical groups such as the azo $\textrm{N=N}$, $\textrm{NO}_2$ and $\textrm{NH}_2$ for the mutagenic class). Specifically, $739/933$ instances containing the subgraph patterns fall into the mutagenic class in the entire dataset, which corroborates that these patterns are sufficient for the ground-truth explanation. Figure~\ref{fig:mutag_motif} describes the detailed distribution of instances with various occurring subgraph patterns. With these occurring subgraph patterns, we can evaluate the explanation performance on Mutag with edge AUC. The evaluation intuition is elaborated in the ~\secref{subsec:settings}.

\begin{figure}[t]
  \centering
  \includegraphics[scale = 0.3]{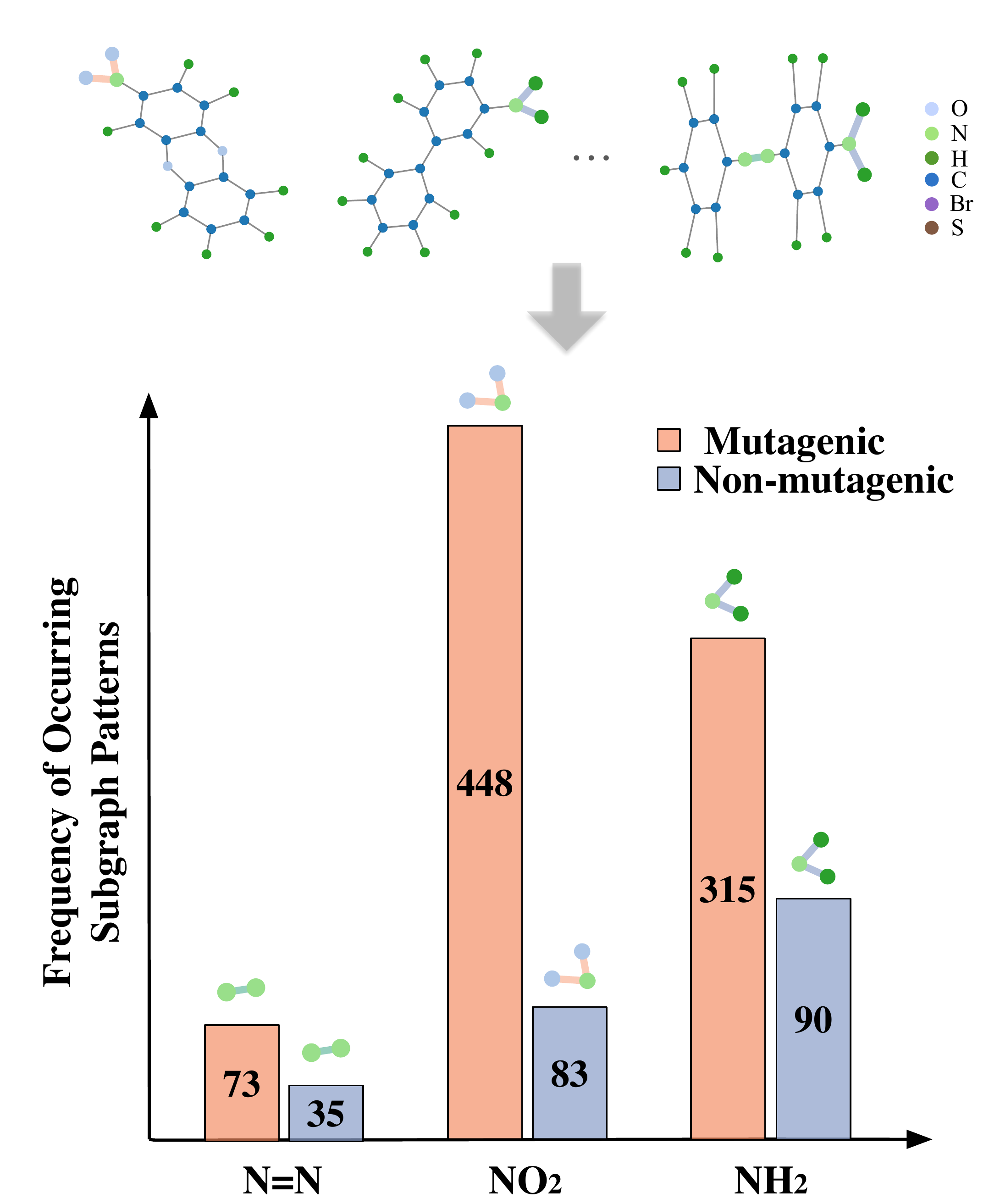}
  \caption{The frequency of occurring subgraph patterns indicates that it is reasonable to treat the labeled motifs/subgraph patterns as the explanation ground truth, i.e., carbon rings with chemical groups such as $\textrm{N=N}$, $\textrm{NO}_2$, and $\textrm{NH}_2$ for the mutagenic class.}
  \vspace{-10pt}
  \label{fig:mutag_motif}
\end{figure}



\begin{table}[!t]
\caption{Explanation Accuracy on Synthetic Datasets ($\%$).}
\vspace{-7pt}

\begin{adjustbox}{center}
\begin{small}
\setlength{\tabcolsep}{0.15em}
\begin{tabular}{ c l || c| c |c |c| c || c |c| c| c |c}
\cmidrule[1pt]{2-12}
& K& \multicolumn{5}{{c}}{BA-SHAPES}  & \multicolumn{5}{{c}}{TREE-CYCLES}\\
&\#of edges &5&6&7&8&9&6&7&8&9&10\\
\cmidrule{2-12}
&{\em OrphicX}&$\textbf{82.4}$&$\textbf{97.1}$&$\textbf{97.1}$&$\textbf{97.1}$&$\textbf{100}$&$\textbf{85.7}$&$\textbf{91.4}$&$\textbf{100}$&$\textbf{100}$&$\textbf{100}$\\
&Gem &$64.7$&$94.1$&$91.2$&$91.2$&$91.2$&$74.3$&$88.6$&$\textbf{100}$&$\textbf{100}$&$\textbf{100}$\\
&GNNExp.~&$67.6$&$67.6$&$82.4$&$88.2$&$85.3$&$20.0$&$54.3$&$74.3$&$88.6$&$97.1$\\
&PGExp.~&$59.5$&$59.5$&$59.5$&$59.5$&$64.3$&$76.2$&$81.5$&$91.3$&$95.4$&$97.1$\\
\cmidrule[1pt]{2-12}
\end{tabular}
\end{small} 
\end{adjustbox}
\label{tab:syn}
\vspace{-20pt}
\end{table} 

\subsection{Empirical Results}
\label{subsec:settings}

{\bf Explanation performance.} We first report the explanation performance for synthetic datasets and real-world datasets. In particular, we evaluate the explanation accuracy under various sparseness constraints (i.e., various $R$ for the real-world datasets and various $K$ for the synthetic datasets). Table~\ref{tab:syn} and Table~\ref{tab:real} report the explanation accuracy of different methods specifically. A smaller number of edges (denoted as $K$) or a smaller value of edge ratio (denoted as $R$) indicates that the explanatory subgraphs are more compact. As observed, {\em OrphicX} consistently outperforms baselines across various sparseness constraints over all datasets. As the model architectures for node and graph classification~\cite{xu2018powerful} tasks are different, the performance corroborates that our framework is model architecture-agnostic (see the model architectures in the Appendix).

Following existing works~\cite{wanyuicml21,Matthew2020}, we also evaluate the {\em Log-odds difference} to illustrate the fidelity of generated explanations in a more statistical view. Log-odds difference describes the resulting change in the pre-trained GNNs' outcome by computing the difference (the initial graph and the explanation subgraph) in log odds. The detailed definition of Log-odds difference is elaborated in Appendix~\ref{appendix-implementation}. Figure~\ref{fig:log-odds} depicts the distributions of log-odds difference over the entire test set for synthetic datasets. We can observe that the log-odds difference of {\em OrphicX} is more concentrated around $0$, which indicates {\em OrphicX} can well capture the most relevant subgraphs towards the predictions by the pre-trained GNNs. As {\em OrphicX} exhibits a similar performance trend on other datasets, we present corresponding evaluation results in the Appendix. 

%
 
\begin{table}[!t]
\caption{Explanation Accuracy on Real-World  Datasets ($\%$).}
\vspace{-10pt}
\begin{adjustbox}{center}
\begin{small}
\setlength{\tabcolsep}{0.15em}
\begin{tabular}{ c l || c |c| c| c| c || c| c| c| c| c}
\cmidrule[1pt]{2-12}
& R& \multicolumn{5}{{c}}{Mutag}  & \multicolumn{5}{{c}}{NCI1}\\
&edge ratio&0.5&0.6&0.7&0.8&0.9&0.5&0.6&0.7&0.8&0.9\\
\cmidrule{2-12}
&{\em OrphicX}&$\textbf{71.4}$&$\textbf{71.2}$&$\textbf{77.2}$&$\textbf{78.8}$&$\textbf{83.2}$&$\textbf{66.9}$&$\textbf{72.7}$&$\textbf{77.1}$&$\textbf{81.3}$&$\textbf{85.4}$\\
&Gem &$66.4$&$67.7$&$71.4$&$76.5$&$81.8$&$61.8$&$68.6$&$70.6$&$74.9$&$83.9$\\
&GNNExp.~&$65.0$&$66.6$&$66.4$&$71.0$&$78.3$&$64.2$&$65.7$&$68.6$&$75.2$&$81.8$\\
&PGExp.~&$59.3$&$58.9$&$65.1$&$70.3$&$74.7$&$57.7$&$60.8$&$65.2$&$69.3$&$71.0$\\
\cmidrule[1pt]{2-12}
\end{tabular}
\end{small} 
\end{adjustbox}
\label{tab:real}
\vspace{-20pt}
\end{table} 

\begin{figure}[!ht]
\vspace{-10pt}%
 \centering
 \vspace{-5pt}%
 \subfloat[BA-shapes]{\includegraphics[scale = 0.25]{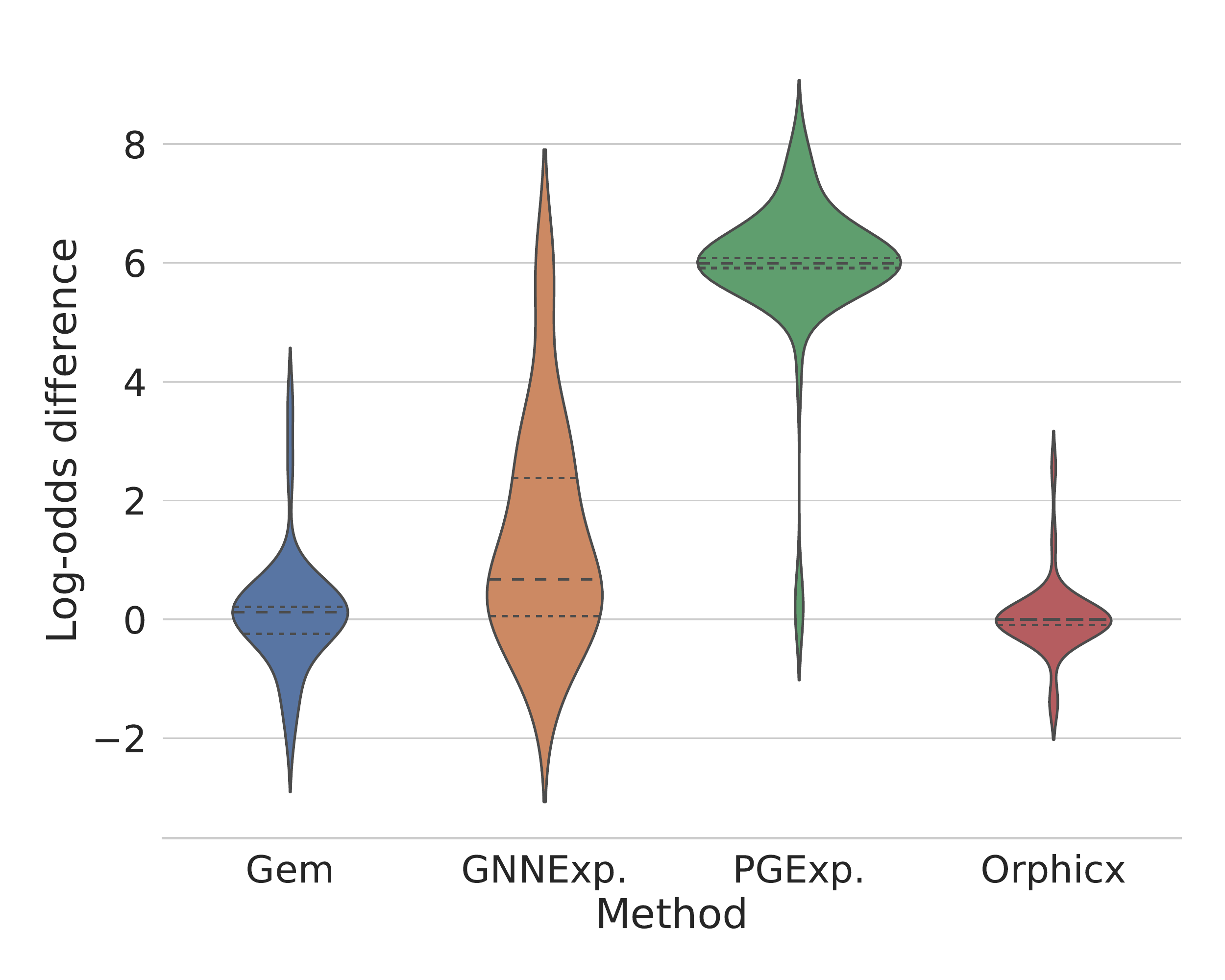}\label{fig:ba-shapes_log_odd}}\\
  \vspace{-3pt}

 \subfloat[Tree-cycles]{\includegraphics[scale = 0.25]{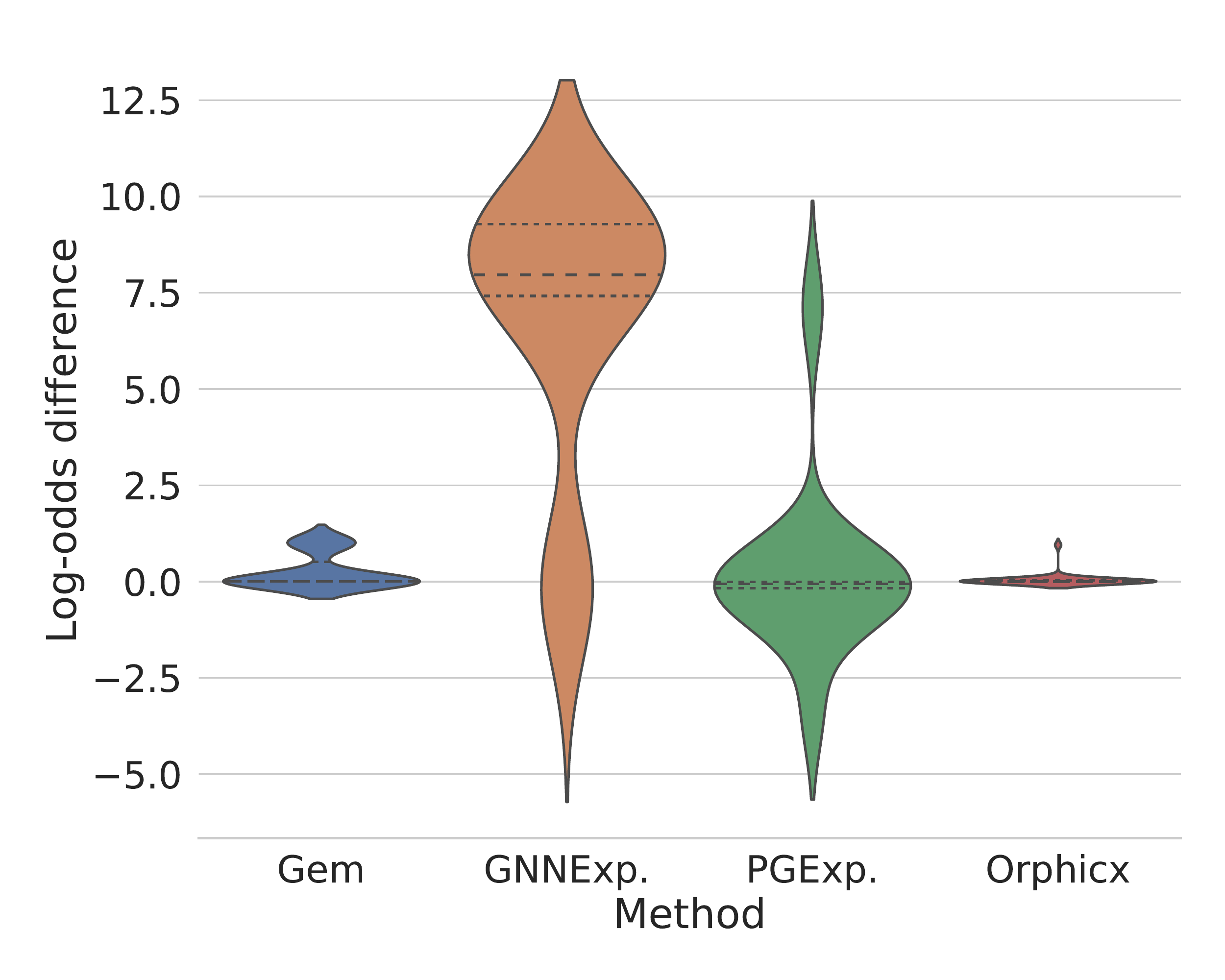}\label{fig:tree-cycle_log_odd}}
 \vspace{-10pt}
 \caption{Explanation Performance with Log-Odds Difference. {\em OrphicX} consistently achieves the best performance overall ({\em denser distribution around $0$ is better}).}%
 \label{fig:log-odds}
 \vspace{-13pt}%
\end{figure}

\begin{table*}[!t]
\caption{Explanation Accuracy with Edge AUC (* means the rounded estimate of $0.9995\pm0.0006$).}
\vspace{-15pt}
\begin{center}
\begin{small}\small\addtolength{\tabcolsep}{-1pt}
\begin{sc}
\setlength{\tabcolsep}{0.40em}
\begin{tabular}{ c| c |c |c| c|c}
\toprule
Datasets & {\em OrphicX} & Gem & GNNExp.~& PGExp.~& ATT\\ 
\midrule
BA-SHAPES &$\mathbf{0.988\pm0.008}$ & $0.597\pm0.001 $ & $0.956\pm0.001$ & $0.924\pm0.042$& $0.815$\\
TREE-CYCLES &$\mathbf{0.988\pm0.001}$ & $0.761\pm0.002  $ & $0.961\pm0.003$ & $0.952\pm0.000$& $0.824$\\
Mutag&$\mathbf{1.000\pm0.001}^*$  & $0.988\pm0.013 $ & $0.998\pm0.001$ & $0.998\pm0.001$& $0.686\pm0.098$\\
\bottomrule
\end{tabular}
\end{sc}
\end{small} 
\end{center}
\vspace{-15pt}
\label{tab:auc}
\end{table*}

\begin{figure*}[t]
  \centering
  \includegraphics[scale = 0.22]{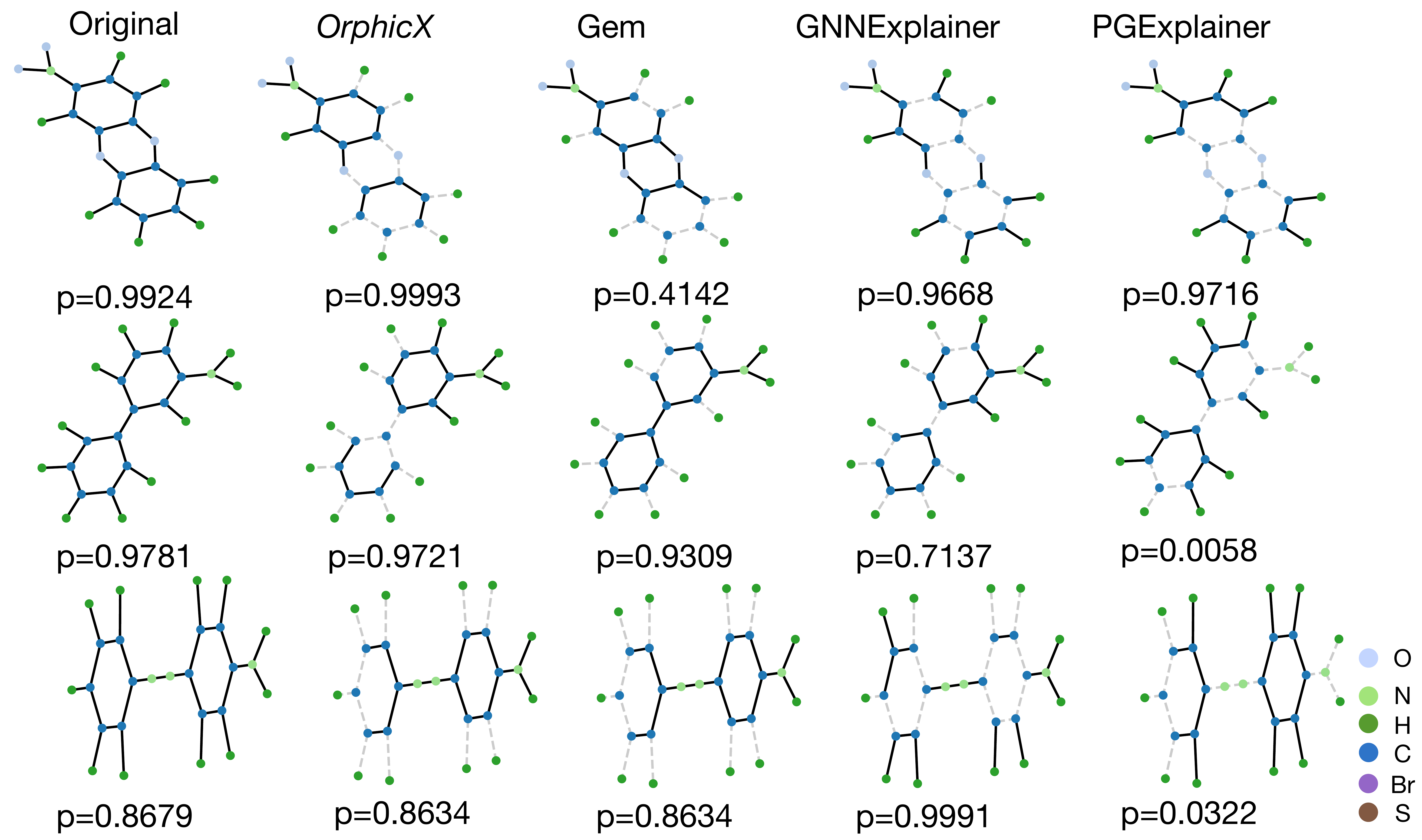}  \vspace{-5pt}
  \caption{Explanation Visualization (MUTAG): $p$ is the corresponding probability of being classified as Mutagenic class by the pre-trained GNN. The graphs in the first column are the target instances to be explained. The solid edges in other columns are identified as `important' by corresponding methods. The closer the probability to that of the target instance, the better the explanation is.}
  \label{fig:graph-visual}
    \vspace{-15pt}
\end{figure*}

For fair comparisons, we also report the explanation fidelity of different methods in terms of edge AUC in Table~\ref{tab:auc}. We follow the experimental settings of GNNExplainer and PGExplainer\footnote{We use PGExp.~and GNNExp.~ to represent PGExplainer and GNNExplainer for simplicity.}, where the explanation problem was formalized as a binary classification of edge. The mean and standard deviation are calculated over $5$ runs. This metric works for the datasets with ground-truth explanations (i.e., the ``house''-structured pattern/motif of BA-shapes and the labeled subgraph patterns in Mutag). The intuition is that a good explanation method assigns higher weights to the edges within the ground-truth subgraphs/motifs. Regarding edge importance, one might naturally consider the self-attention mechanism as a feasible solution. Prior works have shown its performance for model explanations. For clarity, we also report the experimental results of the self-attention mechanism denoted as~\textbf{ATT} in Table~\ref{tab:auc}. The results of synthetic datasets are from GNNExplainer and PGExplainer. For Mutag, we evaluate the subgraph patterns labeled by the domain expert. As might be expected, {\em OrphicX} exhibits its superiority in identifying the most important edges captured by the pre-trained GNNs. We also observe that the prior causality-based approach, Gem, does not perform well evaluating with edge AUC. We conjecture that the explainable subgraph patterns are destroyed due to the distillation process~\cite{wanyuicml21}. Though the generated subgraphs with Gem can well reflect the classification pattern captured by the pre-trained GNN, it degrades the human interpretability of the generated explanations.

{\em Explanation visualization.} Figure~\ref{fig:graph-visual} plots the visualized explanations of different methods. In particular, we focus on the visualization on Mutag, which can reflect the interpretability quantitatively and qualitatively. The first column shows the initial graphs and corresponding probabilities of being classified as ``mutagenic'' class by the pre-trained GNN, while the other columns report the explanation subgraphs. Associated probabilities belonging to the ``mutagenic'' class based on the pre-trained GNN are reported below the subgraphs. Specifically, in the first case (the first row), {\em OrphicX} can identify the essential subgraph pattern --- a complete carbon ring with a $\textrm{NO}_2$ --- leading to its label ( ``mutagenic''). Nevertheless, prior works, particularly Gem, fail to recognize the explainable motif. In the second instance (the second row), {\em OrphicX} can well identify a complete carbon ring with a $\textrm{NH}_2$. At the same time, PGExplainer fails to recognize the $\textrm{NH}_2$, leading to a high probability of being classified into the wrong class --- ``non-mutagenic'' --- by the target GNN, with a probability of $0.9942$. In the third instance (the third row), a complete carbon ring with a $\textrm{N=N}$ is the essential motif, consistent with the criterion from the domain expert. Overall, {\em OrphicX} can identify the explanatory subgraphs that best reflect the predictions of the pre-trained GNN. The visualization of synthetic datasets and more visualization plots on Mutag are provided in Appendix~\ref{sec:appendix_results}.


{\bf Information flow measurements.} To validate Theorem~\ref{thm:info_flow}, we evaluate the information flow of the causal factors ($\Z_c$) and the spurious factors ($\Z_s$) corresponding to the model prediction, respectively. Figure~\ref{fig:mutag_info} in Appendix~\ref{sec:appendix_results} shows that, as desired, the information flow from the causal factors to the model prediction is large while the information flow from the spurious factors to the prediction is small. We also evaluate the prediction performance while adding noise (mean is set as 0) to the causal factors and the spurious factors, respectively. From Table~\ref{tab:noise-auc}, we can observe that adding perturbations to the causal factors degrades the prediction performance of the pre-trained GNN significantly with the increase of the standard deviation of the noise (mean is set as 0) while adding the perturbations on the spurious counterparts does not. These insights, in turn, verify the efficacy of applying the concept of information flow for the causal influence measurements.

\begin{table}[!t]
\caption{Prediction Accuracy of the Pre-trained GNN on Mutag with Various Perturbation (mean is set as 0).}
\vspace{-14pt}

\begin{center}
\begin{small}\small\addtolength{\tabcolsep}{-1pt}
\begin{sc}
\setlength{\tabcolsep}{0.2em}
\begin{tabular}{ c| c |c |c| c|c|c}
\toprule
Perturbation std & 0.0 & 0.3 & 0.5& 0.8& 1.0& 1.3 \\
\midrule
Causal factors&$0.935$ & $0.926 $ & $0.926$ & $0.887$& $0.860$ &$0.826$\\
\midrule
Spurious factors&$0.935$ & $0.936  $ & $0.936$ & $0.935$& $0.934$&$0.926$\\
\bottomrule
\end{tabular}
\end{sc}
\end{small} 
\end{center}
\vspace{-25pt}
\label{tab:noise-auc}
\end{table}

{\bf Ablation studies.} An ablation study for the information flow in the hidden space was performed by removing the causal influence term. From Figure~\ref{fig:mutag_no_causal_info} in Appendix~\ref{sec:appendix_results}, we can observe that without the causal influence term, the causal influence to the model prediction is distributed across all hidden factors. In addition, we also inspect the explanation performance for our framework as an ablation study for the loss function proposed. We empirically prove the need for different forms of regularization leveraged by the {\em OrphicX} loss function. Due to space constraints, the empirical results are provided in the Appendix.


\section{Related Work}

We focus on the discussions on causality-based interpretation methods. Other prior works, including GNNExplainer~\cite{ying2019gnnexplainer}, PGExplainer~\cite{luo2020parameterized}, PGM-Explainer~\cite{vu2020pgm}, SubgraphX~\cite{yuan2021explainability}, GraphMask~\cite{schlichtkrull2021interpreting}, XGNN~\cite{yuan2020xgnn} and others~\cite{pope2019explainability} are provided in Appendix~\ref{sec:other-related}. 

Explanation essentially seeks the answers to the questions of ``what if'' and ``why,'' which are intrinsically causal. Causality, therefore, has been a plausible language for answering such questions~\cite{Matthew2020,wanyuicml21}. There are several viable formalisms of causality, such as structural causal models~\cite{pearl2009causality,Matthew2020}, Granger causality~\cite{granger1969investigating,wanyuicml21}, and causal Bayesian networks~\cite{pearl2009causality}. While most existing works are designed for explaining conventional neural networks on image domain, Gem~\cite{wanyuicml21} falls into the research line of explaining graph-structural data. Specifically, Gem framed the explanation task for GNNs as a causal learning task and proposed a causal explanation model that can learn to generate compact subgraphs towards its prediction. Fundamentally, this approach monitored the response of the target GNN by perturbing the input aspects in the data space and naturally impelled the independent assumption of the explained features. Due to the interdependence property of graph-structured data and the non-linear transformation of GNNs, we argue that this assumption may reduce the efficacy and optimality of the explanation performance. Different from prior works, we quantify the causal attribution of the data aspects in the latent space, and we do not have the independent assumption of the explained features, as {\em OrphicX} is designed to generate the explanations as a whole.

\textbf{Graph information bottleneck.} Our work is somewhat related to the work of information bottleneck for subgraph recognition~\cite{yu2021graph} but different in terms of the problem and the goals. GIB-SR~\cite{yu2021graph} seeks to recognize maximally informative yet compressed subgraph given the input graph and its properties (e.g, ground truth label). On the contrary, our framework is about generating explanations to unveil the inner working of GNNs, which seeks to understand the behavior of the target model (the prediction results) rather than the ground truth labels. More concretely, the model explanation is to analyze models rather than data~\cite{molnar2020interpretable}. Moreover, our objective maximizes the causal information flowing from the latent features to the model predictions.


\section{Conclusion}
\label{sec:conclusion}

In this paper, we propose {\em OrphicX}, a framework for generating causal, compact, and faithful explanations for any graph neural networks. Our findings remain consistent across datasets and various graph learning tasks. Our analysis suggests that {\em OrphicX} can identify the causal semantics in the latent space of graphs via maximizing the information flow measurements. In addition, {\em OrphicX} enjoys several advantages over many powerful explanation methods: it is model-agnostic, and it does not require the knowledge of the internal structure of the target GNN, nor rely on the linear-independence assumption of the explained features. We show that causal interpretability via isolating the causal factors in the latent space offers a promising tool for explaining GNNs and mining patterns in subgraphs of graph inputs. 

Explainability will promote transparency, trust, and fairness in society. It can be very helpful for graphs, including but not limited to molecular graphs, for example, visual scene graph --- a graph-structured data where nodes are objects in the scene and edges are relationships between objects. Explainability can identify subgraphs relevant to a given classification, e.g., identify a scene as being indoor. In the future, additional user studies should confirm to what extent explanations in other domains (e.g., visual scene graph) provided by our {\em OrphicX} align with the needs and requirements of practitioners in real-world settings.

{\em Potential negative impact.} The privacy risks of model explanations have been empirically characterized for deep neural networks for non-relational data (with respect to graph-structured data)~\cite{shokri2021privacy}. We conjecture that the generated explanation for GNNs may also expose private information of the training data. This will pose risks for deploying GNN-based AI systems across various domains that value model explainability and privacy the most, such as finance and healthcare. 

\section{Acknowledgement}
The authors thank the reviewers/AC for the constructive comments to improve the paper. This project is supported by the Internal Research Fund at The Hong Kong Polytechnic University P0035763. HW is partially supported by NSF Grant IIS-2127918 and an Amazon Faculty Research Award.

{\small
\bibliographystyle{ieee_fullname}
\bibliography{neurips_2021}
}
\clearpage
\appendix
\section{Appendix}

\subsection{Derivation of the Estimator in Theorem 2.1 and \eqnref{eq:zcyzc}}
\label{sec:appendix-proof}
\begingroup\makeatletter\def\f@size{7.5}\check@mathfonts
\begin{align*}
I(\Z_c\rightarrow y)& = \int_{\Z_c}  P(\Z_c)\Big( \sum_y P(y|do(\Z_c)) \log P(y|do(\Z_c)) ) \Big) d\Z_c \\
&- \sum_y \int_{\Z_c} P(\Z_c)P(y|do(\Z_c)) d\Z_c\cdot\\
&\log \Big(\int_{\Z_c} P(\Z_c)P(y|do(\Z_c)) d\Z_c\Big) \\
= & \frac{1}{N_c} \sum_{i=1}^{N_c} \sum_y \Big(\frac{1}{N_x N_s N_z} \sum\limits_{k=1}^{N_x}\sum\limits_{j=1}^{N_s}\sum\limits_{n=1}^{N_z}P(y|\A^{(ikjn)}, \X^{(k)})\Big)\cdot\\
&\log \Big(\frac{1}{N_x N_s N_z} \sum\limits_{k=1}^{N_x}\sum\limits_{j=1}^{N_s}\sum\limits_{n=1}^{N_z}P(y|\A^{(ikjn)}, \X^{(k)})\Big) \\
&- \sum_y \Big(\frac{1}{N_c N_x N_s N_z} \sum\limits_{i=1}^{N_c} \sum\limits_{k=1}^{N_x}\sum\limits_{j=1}^{N_s}\sum\limits_{n=1}^{N_z}P(y|\A^{(ikjn)}, \X^{(k)})\Big)\cdot\\
&\log\Big(\frac{1}{N_c N_x N_s N_z} \sum\limits_{i=1}^{N_c} \sum\limits_{k=1}^{N_x}\sum\limits_{j=1}^{N_s}\sum\limits_{n=1}^{N_z}P(y|\A^{(ikjn)}, \X^{(k)})\Big) \\
= & \frac{1}{N_c N_x N_s N_z} \Big[ \sum_{i=1}^{N_c} \sum_y \Big( \sum\limits_{k=1}^{N_x}\sum\limits_{j=1}^{N_s}\sum\limits_{n=1}^{N_z}P(y|\A^{(ikjn)}, \X^{(k)})\Big)\cdot\\
&\log \Big(\frac{1}{N_x N_s N_z} \sum\limits_{k=1}^{N_x}\sum\limits_{j=1}^{N_s}\sum\limits_{n=1}^{N_z}P(y|\A^{(ikjn)}, \X^{(k)})\Big) \\
&- \sum_y \Big( \sum\limits_{i=1}^{N_c} \sum\limits_{k=1}^{N_x}\sum\limits_{j=1}^{N_s}\sum\limits_{n=1}^{N_z}P(y|\A^{(ikjn)}, \X^{(k)})\Big)\cdot\\
& \log \Big(\frac{1}{N_c N_x N_s N_z} \sum\limits_{i=1}^{N_c} \sum\limits_{k=1}^{N_x}\sum\limits_{j=1}^{N_s}\sum\limits_{n=1}^{N_z}P(y|\A^{(ikjn)}, \X^{(k)})\Big) \Big]
\end{align*}
\endgroup

\subsection{Further Implementation Details}
\label{appendix-implementation}

{\bf Datasets.} BA-shapes was created with a base Barabasi-Albert (BA) graph containing $300$ nodes and $80$ five-node ``house''-structured network motifs. Tree-cycles were built with a base $8$-level balanced binary tree and $80$ six-node cycle motifs. Mutag~\cite{debnath1991structure} and NCI1~\cite{wale2008comparison} are for graph classification tasks. Specifically, Mutag contains 4337 molecule graphs, where nodes represent atoms, and edges denote chemical bonds. It contains the non-mutagenic and mutagenic class, indicating the mutagenic effects on Gram-negative bacterium Salmonella typhimurium. NCI1 consists of 4110 instances; each chemical compound screened for activity against non-small cell lung cancer or ovarian cancer cell lines. The statistics of four datasets are presented in Table~\ref{tab:data_sta}. Note that, we report the average number of nodes and the average number of edges over all the graphs for the real-world datasets.

\begin{table}[!ht]
\caption{Data Statistics of Four Datasets.}
\begin{center}
\begin{small}\small\addtolength{\tabcolsep}{-3pt}
\begin{sc}
\setlength{\tabcolsep}{0.5em}
\begin{tabular}{ c| c |c |c| c}
\toprule
Datasets & BA-shapes & Tree-cycles & Mutag & NCI1\\ 
\midrule
\#graphs&$1$ & $1$ & $4,337$ & $4,110$\\
\#nodes&$700$&$871$&$29$&$30$\\
\#edges&$4,110$&$1,950$&$30$&$32$\\
\#labels&$4$&$2$&$2$&$2$\\
\bottomrule
\end{tabular}
\end{sc}
\end{small}	
\end{center}
\label{tab:data_sta}
\end{table}



\begin{table}[!ht]
\caption{Model Accuracy of Four Datasets ($\%$).}
\begin{center}
\begin{small}\small\addtolength{\tabcolsep}{-5pt}
\begin{sc}
\setlength{\tabcolsep}{0.07em}
\begin{tabular}{ c| c |c |c| c}
\toprule
Datasets & BA-shapes & Tree-cycles & Mutag & NCI1\\ 
\midrule
Accuracy&$94.1$ & $97.1$ & $88.5$ & $78.6$\\
\bottomrule
\end{tabular}
\end{sc}
\end{small}	
\end{center}
\label{tab:data_acc}
\end{table}

{\bf Model architectures.} For classification architectures, we use the same setting as prior works~\cite{wanyuicml21,ying2019gnnexplainer}. Specifically, for node classification, we apply three layers of GCNs with output dimensions equal to $20$ and perform concatenation to the output of three layers, followed by a linear transformation to obtain the node label. For graph classification, we employ three layers of GCNs with dimensions of $20$ and perform global max-pooling to obtain the graph representations. Then a linear transformation layer is applied to obtain the graph label. Figure~\ref{fig:model-arch} (a) and ~\ref{fig:model-arch} (b) are the model architectures for node classification and graph classification, receptively. 

Figure~\ref{fig:model-arch} (c) depicts the model architecture of {\em OphicX} for generating explanations. For the inference network, we applied a three-layer GCN with output dimensions $32$, $32$, and $16$. The generative model is equipped with a two-layer MLP and an inner product decoder. We trained the explainers using the Adam optimizer~\cite{kingma2014adam} with a learning rate of $0.003$ for $300$epochs. Table~\ref{tab:stats} shows the detailed data splitting for model training, testing, and validation. Note that both classification models and our explanation models use the same data splitting. See Table~\ref{tab:hyper} for our hyperparameter search space. Table~\ref{tab:data_acc} reports the model accuracy on four datasets, which indicates that the models to be explained are performed reasonably well. Unless otherwise stated, all models, including GNN classification models and our explanation model, are implemented using PyTorch~\footnote{https://pytorch.org} and trained with Adam optimizer.

\begin{table}[!ht]
\caption{Data Splitting for Four Datasets.}
\begin{center}
\begin{small}\small\addtolength{\tabcolsep}{-5pt}
\begin{sc}
\setlength{\tabcolsep}{0.07em}
\begin{tabular}{ c| c |c |c}
\toprule
Datasets & \#of Training & \#of Testing & \#of Validation\\ 
\midrule
Ba-shapes&$300$ & $50$ & $50$\\
Tree-cycles&$270$ & $45$ & $45$\\
Mutag&$3,468$ & $434$ & $434$\\
NCI1&$3,031$ & $410$ & $411$\\
\bottomrule
\end{tabular}
\end{sc}
\end{small}	
\end{center}
\label{tab:stats}
\end{table}

\begin{figure}[ht!]
  \centering
  \includegraphics[scale = 0.06]{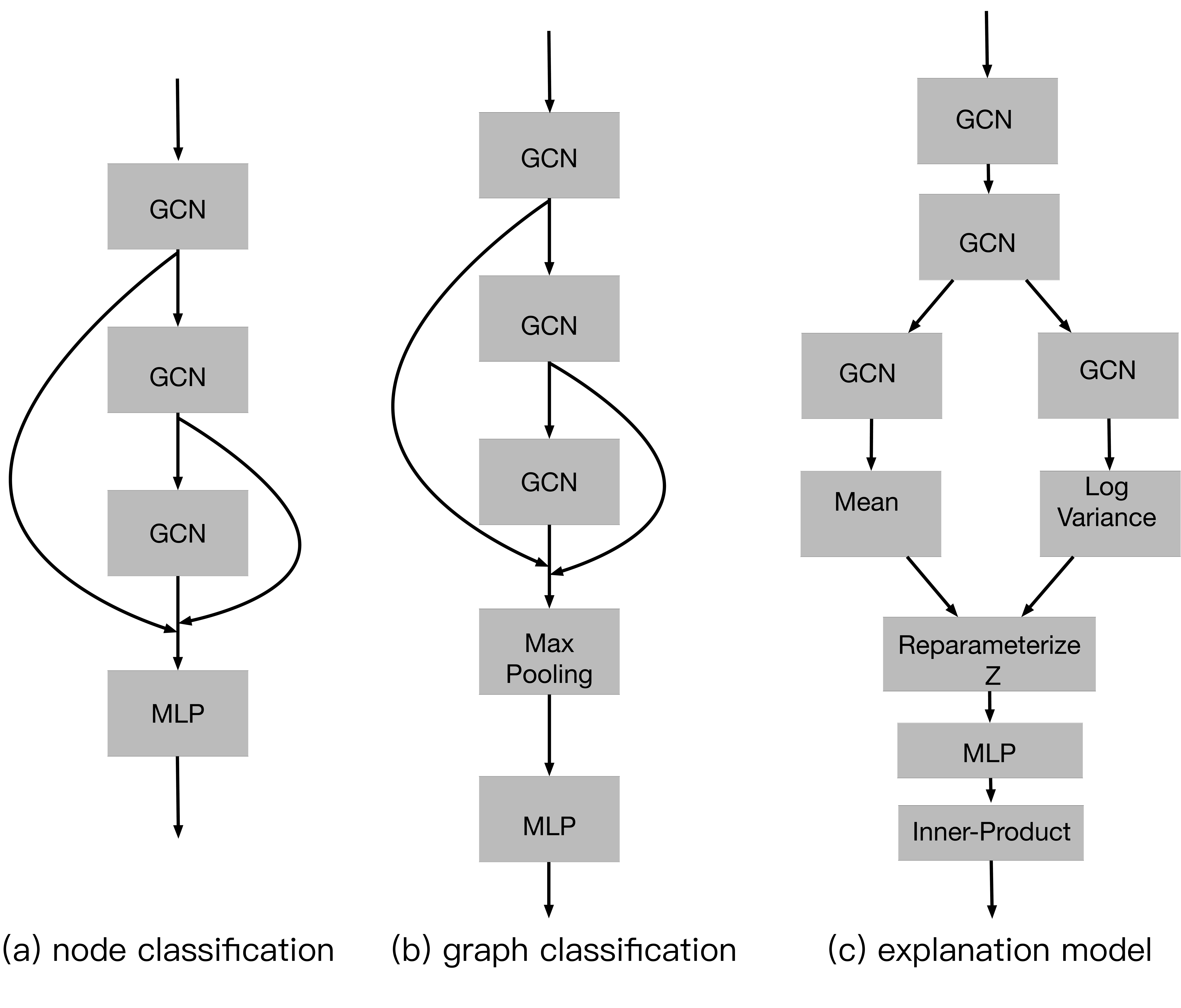}
  \caption{Model architectures.}
  \label{fig:model-arch}
\end{figure}

\begin{table}[!ht]
\caption{Hyperparameters and Ranges}
\begin{center}
\begin{small}\small\addtolength{\tabcolsep}{-3pt}
\begin{sc}
\setlength{\tabcolsep}{0.07em}
\begin{tabular}{ c  c}
\toprule
Hyperparameter& Range \\ 
\midrule
Causal dimension $\D_c$ &$\{1,2,3,\cdots,8\}$\\
negative ELBO $\lambda_1$ & $\{0, 0.01, 0.02, 0.05, 0.1, 0.2, 0.5,1 \}$\\
Sparsity $\lambda_2$&$\{0, 0.01, 0.02, 0.05, 0.1, 0.2, 0.5,1\}$\\
fidelity $\lambda_3$&$\{0, 0.01, 0.02, 0.05, 0.1, 0.2, 0.5,1 \}$\\
\bottomrule
\end{tabular}
\end{sc}
\end{small}	
\end{center}
\label{tab:hyper}
\end{table}

{\bf Negative ELBO term.} The negative ELBO term is defined as~\eqnref{eq:elbo}:

\begin{equation}\label{eq:elbo}
\mathcal{L}_{\mathbf{VGAE}}=\mathbb{E}_{q(\Z|\X, \A)} [\log p(\A|\Z)]-\mathbf{KL}[q(\Z|\X, \A)\parallel p(\Z)],
\end{equation}
where $\mathbf{KL}[q(\cdot)\parallel p(\cdot)]$ is the Kullback-Leibler divergence between $q(\cdot)$ and $p(\cdot)$. The Gaussian prior is $p(\Z)=\prod_{i} p(\z_i)=\prod_{i} \mathcal{N}(\z_i|0,\mathbf{1})$. We follow the reparameterization trick in~\cite{kipf2016variational} for training.

{\bf Log-odds difference.} We measure the resulting change in the pre-trained GNNs' outcome by computing the difference in log odds and investigate the distributions over the entire test set. The log-odds difference is formulated as:
\begin{equation}\label{eq:log}
\Delta\textrm{log-odds}=\textrm{log-odds}\left(f(G)\right)-\textrm{log-odds}\left(f(G_c)\right)
\end{equation}
where $\textrm{log-odds}(p)=\mathbf{log}\left(\frac{p}{1-p}\right)$, and $f(G)$ and $f(G_c)$ are the outputs of the pre-trained GNN. Figure~\ref{fig:log-odds-real} depicts the distributions of log-odds difference over the entire test set for the real-world datasets. 

\subsection{More Experimental Results}
\label{sec:appendix_results}

 {\bf Log-odds difference on the real-world datasets.} Figure~\ref{fig:log-odds-real} depicts the distributions of log-odds difference over the entire test set for the real-world datasets. We can observe that the log-odds difference of {\em OrphicX} is more concentrated around $0$, which indicates {\em OrphicX} can well capture the most relevant subgraphs towards the predictions by the pre-trained GNNs. 

\begin{figure}[!ht]
 \centering
 \subfloat[MUTAG]{\includegraphics[scale = 0.25]{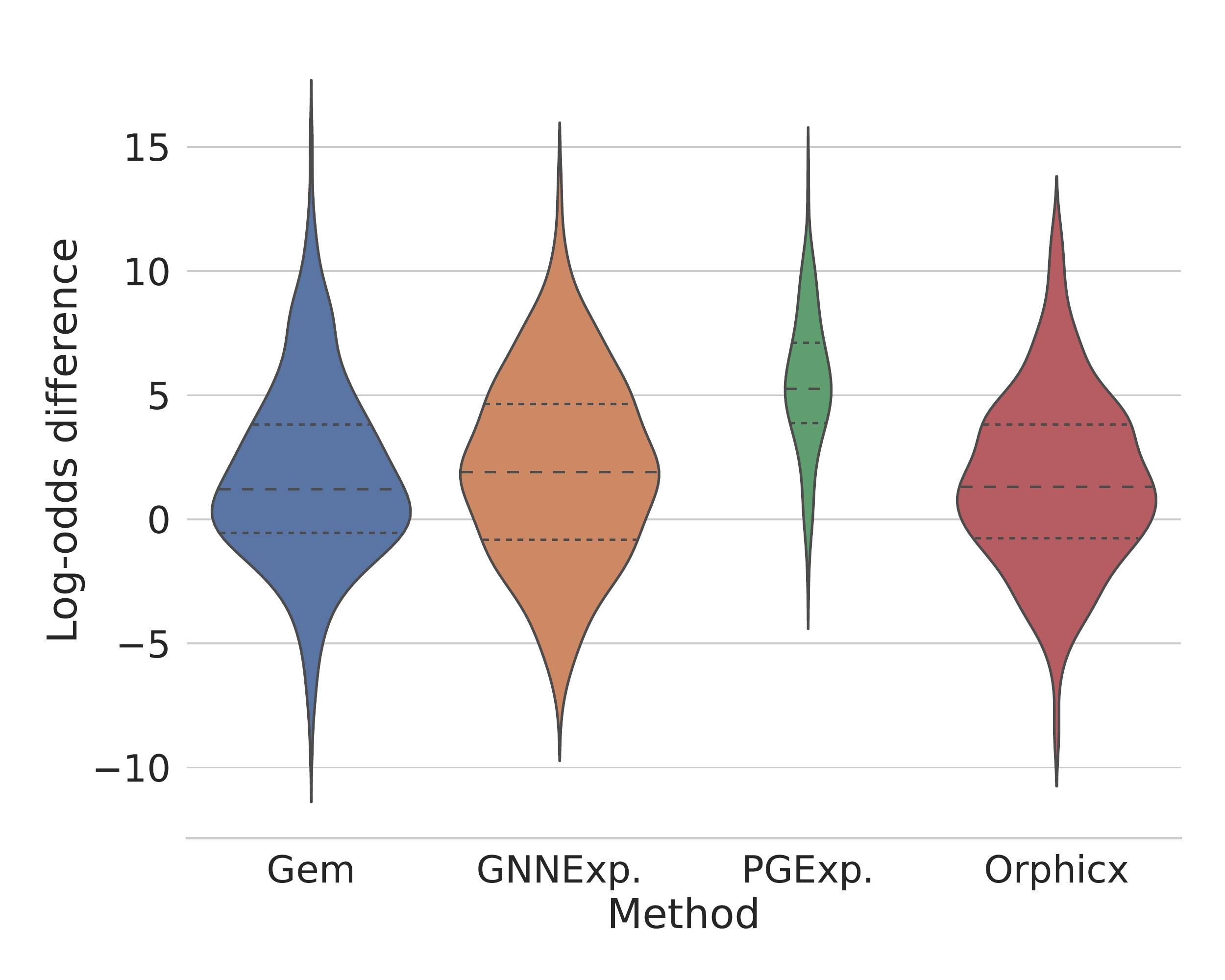}\label{fig:mutag_log_odd_0.7}}\\
 \subfloat[NCI1]{\includegraphics[scale = 0.25]{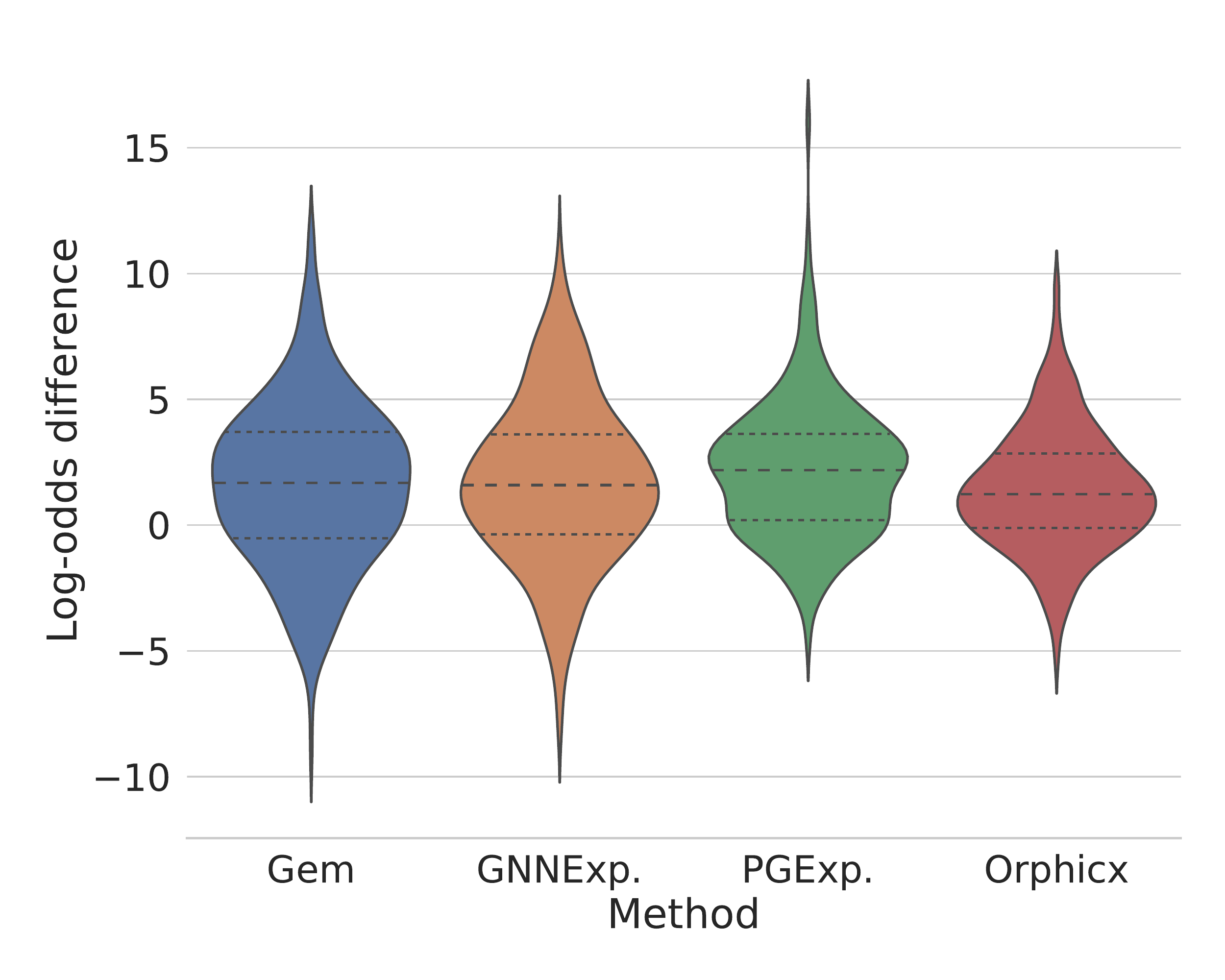}\label{fig:nci1_log_odd_0.6}}
 \caption{Explanation Performance with Log-Odds Difference. {\em OrphicX} consistently achieves the best performance overall ({\em denser distribution around $0$ is better}).}%
 \label{fig:log-odds-real}
\end{figure}

{\bf More visualization results.} Figure~\ref{fig:node-visual} plots the visualized explanations of different methods on BA-shapes. The ``house'' in green is the ground-truth motif that determines the node labels. The red node is the target node to be explained. By looking at the explanations for a target node (the instance on the left side), shown in Figure~\ref{fig:node-visual}, {\em OrphicX} can successfully identify the ``house'' motif that explains the node label (``middle-node'' in red), when $K=6$, while GNNExplainer wrongly attributes the prediction to a node (in orange) that is out of the ``house'' motif. For the right one, {\em OrphicX} consistently performs well, while Gem and GNNExplainer both fail when $K=6$. Figure~\ref{fig:graph-appendix} plots more visualized explanations of different methods on Mutag. 

\begin{figure*}[t]
  \centering
  \includegraphics[scale = 0.45]{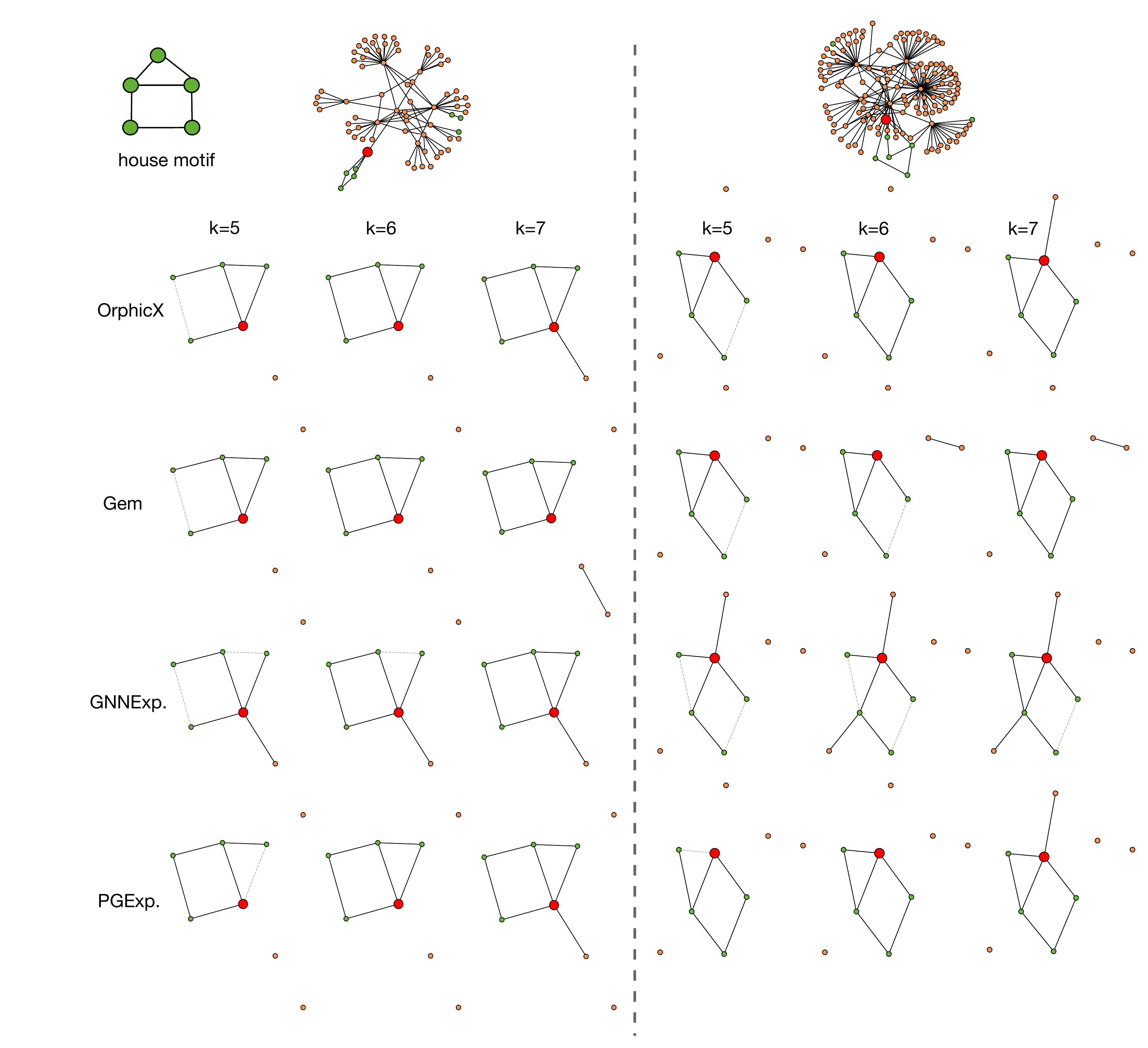}
  \caption{Explanation comparisons on BA-shapes. The ``house'' in green is the ground-truth motif that determines the node labels. The red node is the target node to be explained (better seen in color).}
  \label{fig:node-visual}
\end{figure*}

\begin{figure*}[t]
  \centering
  \includegraphics[scale = 0.24]{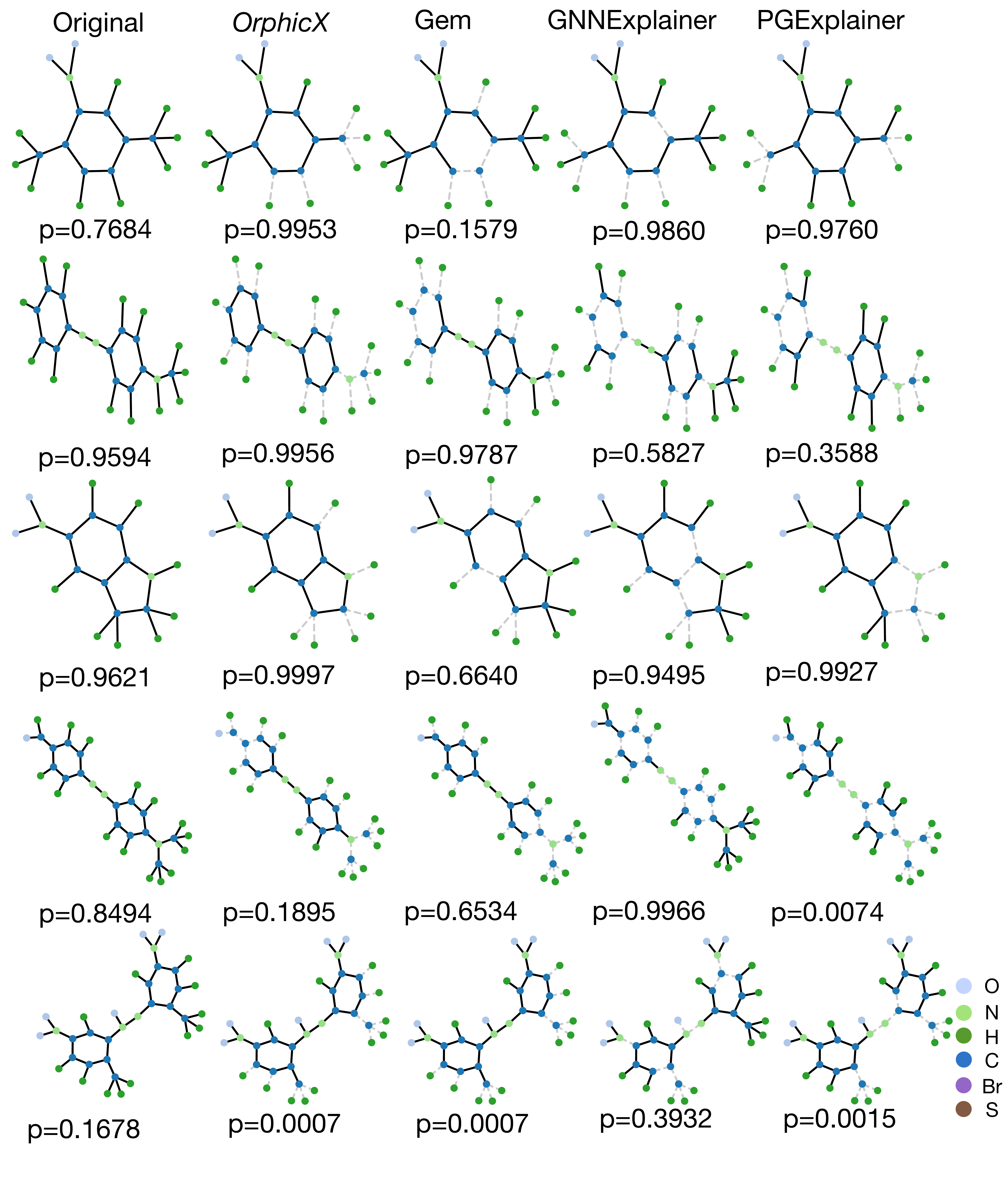}
  \caption{Explanation Visualization (MUTAG): $p$ is the corresponding probability of being classified as Mutagenic class by the pre-trained GNN. The graphs in the first column are the target instances to be explained. The solid edges in other columns are identified as `important' by corresponding methods. The closer the probability to that of the target instance, the better the explanation is.}
  \label{fig:graph-appendix}
\end{figure*}

\begin{table}[!t]
\caption{Causal Evaluation ($\%$).}
\begin{adjustbox}{center}
\begin{small}
\setlength{\tabcolsep}{0.4em}
\begin{tabular}{ c l || c c c  | c c c }
\cmidrule[1pt]{2-8}
& & \multicolumn{3}{{c}}{Mutag}  & \multicolumn{3}{{c}}{NCI1}\\
\cmidrule{2-8}
&R (edge ratio)&0.7&0.8&0.9&0.7&0.8&0.9\\
\cmidrule{2-8}
&original&$77.2$&$78.8$&$83.2$&$77.1$&$81.3$&$85.4$\\
\cmidrule{2-8}
&deconfounder&$67.1$  & $71.5$ & $81.5$&$71.6$  & $79.2$ & $87.3$\\
\cmidrule[1pt]{2-8}
\end{tabular}
\end{small} 
\end{adjustbox}
\label{tab:causal}
\vspace{-23pt}
\end{table} 

{\bf Causal evaluation.} To further verify that the generated explanations are causal and therefore robust to distribution shift in the confounder (i.e., the node attributes $\X$), we construct harder versions of both datasets. Specifically, we use k-means (k=2) to split the dataset into two clusters according to the node attributes. In Mutag, we use the cluster with 3671 graph instances for explainer training and validation; we evaluate the explaining accuracy of the trained explainer on the other cluster with 665 instances. In NCI1, we use the cluster with 3197 graph instances to train an explainer, in which the training set contains 2558 instances and the validation set contains 639 instances; the explaining accuracy is evaluated with the other cluster with 906 instances. See Table~\ref{tab:causal} for details. We can observe that our approach is indeed robust to the distribution shift in the confounder.

{\bf Information flow measurements.} To validate Theorem~\ref{thm:info_flow}, we evaluate the information flow of the causal factors ($\Z_c=\Z[1:3]$) and the spurious factors ($\Z_s=\Z[4:16]$) corresponding to the model prediction, respectively. Figure~\ref{fig:mutag_info} shows that, as desired, the information flow from the causal factors to the model prediction is large while the information flow from the spurious factors to the prediction is small.

\begin{figure}[!ht]
 \centering
 \subfloat[MUTAG]{\includegraphics[scale = 0.4]{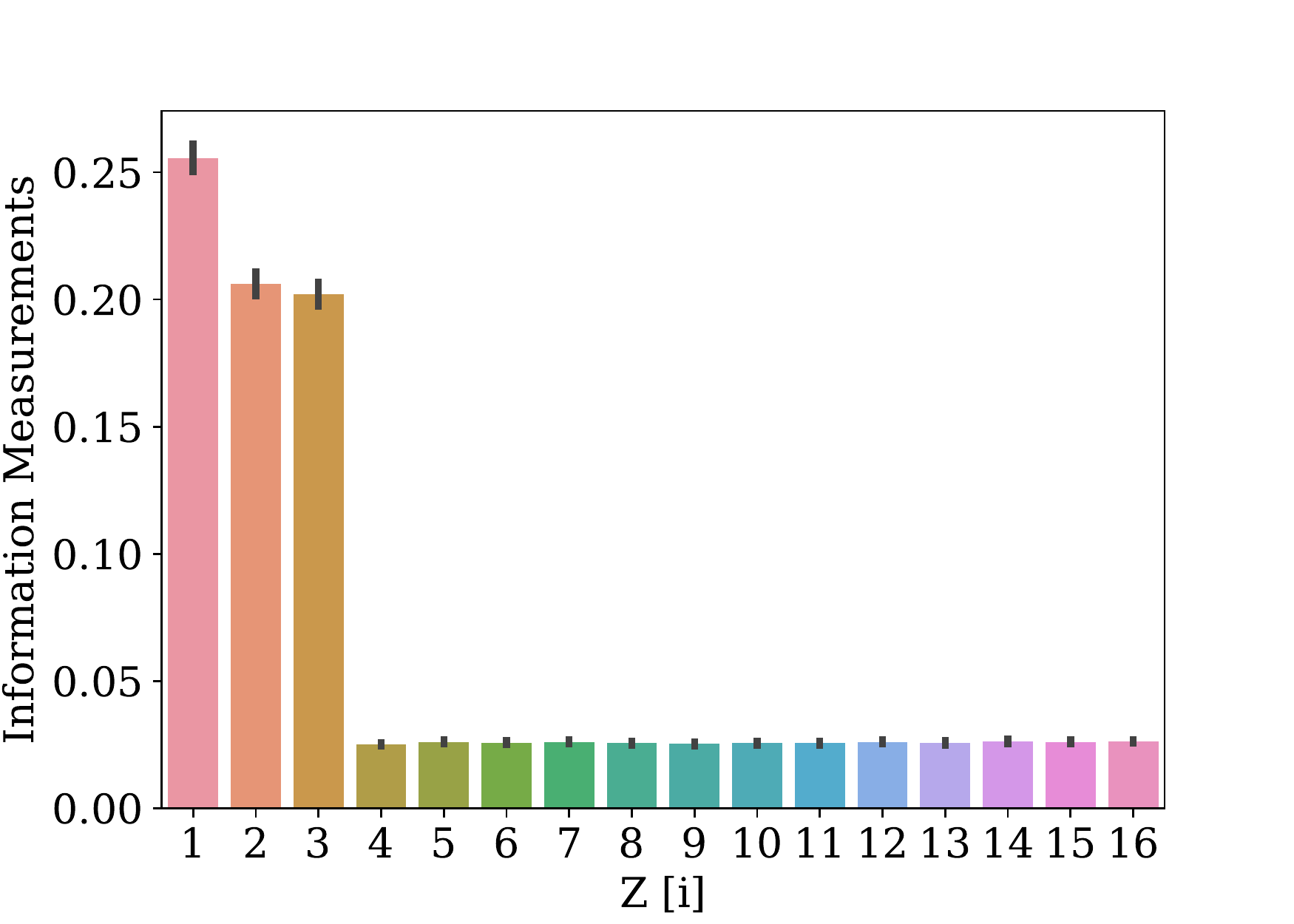}\label{fig:mutag_info}}\\
  \subfloat[MUTAG w/o Causal Term]{\includegraphics[scale = 0.4]{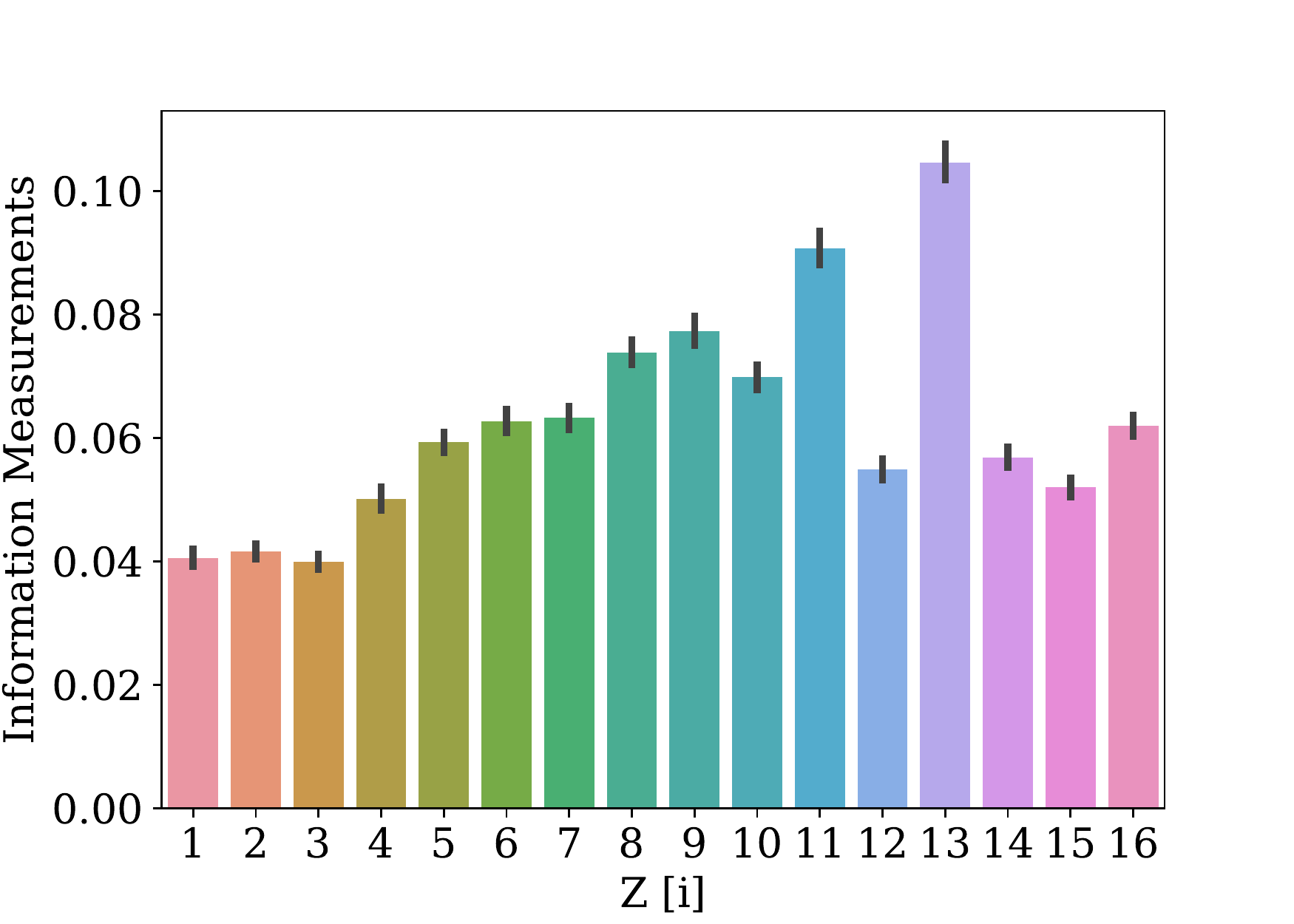}\label{fig:mutag_no_causal_info}}
 \caption{Information Flow Measurements.~\figref{fig:mutag_info} reports the information flow measurements in the hidden space, where $i$ denotes the $i$th dimension.~\figref{fig:mutag_no_causal_info} reports the ones while the causal influence term was removed from the loss function. }%
 \label{fig:info}
\end{figure}

{\bf Ablation study.} We inspect the explanation performance for our framework as an ablation study for the loss function proposed. We empirically prove the need for different forms of regularization leveraged by the {\em OrphicX} loss function. In particular, we compute the average explanation accuracy of $3$ runs. Table~\ref{tab:abla} shows the explanation accuracy of removing a particular regularization term for Mutag and NCI1, respectively. We observe considerable performance gains from introducing the VGAE ELBO term, sparsity, and fidelity penalty. In summary, these results empirically motivate the need for different forms of regularization leveraged by the {\em OrphicX} loss function.

\begin{table}[!ht]
\caption{Ablation Studies for Different Regularization Terms ($\%$).}
\begin{center}
\begin{small}\small\addtolength{\tabcolsep}{-5pt}
\begin{sc}
\setlength{\tabcolsep}{0.03em}
\begin{tabular}{ c| c c c c| c c}
\toprule
Type& Causal & ELBO & Sparsity & Fidelity& Mutag & NCI1 \\ 
& Influence & &  &  & \\ 

\midrule
{\em OrphicX} & \checkmark & \checkmark & \checkmark & \checkmark &$\mathbf{0.854}$ & $\mathbf{0.832}$ \\
			A & \checkmark & \checkmark & \checkmark &  &$0.829$ &$0.633$ \\
		
			B & \checkmark & \checkmark &  & \checkmark & $0.804$& $0.824$\\
			C & \checkmark &  & \checkmark  &\checkmark  & $0.594$& $0.633$\\
\bottomrule
\end{tabular}
\end{sc}
\end{small}	
\end{center}
\label{tab:abla}
\end{table}

{\bf Efficiency evaluation.} {\em OrphicX}, Gem, and PGExplainer can explain unseen instances in the inductive setting. We measure the average inference time for these methods. As GNNExplainer explains an instance at a time, we measure its average time cost per explanation for comparisons. As reported in Table~\ref{tab:inference}, we can conclude that the learning-based explainers such as {\em OrphicX}, Gem, and PGExplaienr are more efficient than GNNExplainer. These experiments were performed on an NVIDIA GTX 1080 Ti GPU with an Intel Core i7-8700K processor.

\begin{table}[!t]
\caption{Explanation Time of Different Methods (Per Instance (ms)).}
\begin{center}
\begin{small}\small\addtolength{\tabcolsep}{-8pt}
\begin{sc}
\setlength{\tabcolsep}{0.07em}
\begin{tabular}{ c| c |c |c| c}
\toprule
Datasets & BA-shapes & Tree-cycles & Mutag & NCI1\\ 
\midrule
{\em OrphicX}&$0.61$ & $2.31$ & $0.01$ & $0.02$\\
Gem&$0.67$ & $0.50$ & $0.05$ & $0.03$\\
GNNExplainer&$260.2$ & $206.5$ & $253.2$ & $262.4$\\
PGExplainer&$6.9$ & $6.5$ & $5.5$ & $5.4$\\
\bottomrule
\end{tabular}
\end{sc}
\end{small} 
\end{center}
\label{tab:inference}
\end{table}

\subsection {More related work on GNN interpretation}%
\label{sec:other-related}

Several recent works have been proposed to provide explanations for GNNs, in which the most important features (e.g., nodes or edges or subgraphs) of an input graph are selected as the explanation to the model's outcome. In essence, most of these methods are designed for generating input-dependent explanations. GNNExplainer~\cite{ying2019gnnexplainer} searches for soft masks for edges and node features to explain the predictions via mask optimization.~\cite{pope2019explainability} extended explainability methods designed for CNNs to GNNs. PGM-Explainer~\cite{vu2020pgm} adopts a probabilistic graphical model and explores the dependencies of the explained features in the form of conditional probability. SubgraphX explores the subgraphs with Monte Carlo tree search and evaluates the importance of the subgraphs with Shapley values~\cite{yuan2021explainability}. In general, these methods explain each instance individually and can not generalize to the unseen graphs, thereby lacking a global view of the target model.  

A recent study has shown that separate optimization for each instance induces hindsight bias and compromises faithfulness~\cite{schlichtkrull2021interpreting}. To this end, PGExplainer~\cite{luo2020parameterized} was proposed to learn a mask predictor to obtain edge masks for providing instance explanations. XGNN~\cite{yuan2020xgnn} was proposed to investigate graph patterns that lead to a specific class. GraphMask~\cite{schlichtkrull2021interpreting} is specifically designed for GNN-based natural language processing tasks, where it learns an edge mask for each internal layer of the learning model. Both these approaches require access to the process by which the target model produces its predictions. As all the edges in the dataset share the same predictor, they might be able to provide a global understanding of the target GNNs. Our work falls into this line of research, as our objective is to learn an explainer that can generate compact subgraph structures contributing to the predictions for any input instances. Different from existing works, we seek faithful explanations from the language of causality~\cite{pearl2009causality}. 



\end{document}